\documentclass{article}

    \PassOptionsToPackage{numbers, compress}{natbib}
\usepackage[preprint]{neurips_2026}

\usepackage[utf8]{inputenc} 
\usepackage[T1]{fontenc}    
\usepackage{hyperref}       
\usepackage{url}            
\usepackage{booktabs}       
\usepackage{amsfonts}       
\usepackage{nicefrac}       
\usepackage{microtype}      
\usepackage{xcolor}         
\usepackage{adjustbox}
\usepackage{array}
\usepackage{graphicx}
\usepackage{multirow}
\usepackage{pifont}
\usepackage{algorithm}
\usepackage{algpseudocode}
\usepackage[normalem]{ulem}
\usepackage{amsmath}
\usepackage{amsthm}

\usepackage{wrapfig}
\usepackage{colortbl}
\usepackage{cuted}
\usepackage{subcaption}
\usepackage{tabularx}
\usepackage{placeins}

\title{MoGe-3: Fine-Detail Monocular Geometry Estimation with Self-Guided Sparse Volumetric Refinement}

\author{%
    Lingyu Kong$^{1,3}$\thanks{These authors contributed equally to this work.}\ \ \thanks{Work done during internship at Microsoft Research} 
    \quad 
    Ruicheng Li$^{1,3}$\footnotemark[1]\ \ \footnotemark[2]
    \quad 
    Ruicheng Wang$^{2,3}$\footnotemark[1]\ \ \footnotemark[2]
    \quad 
    Sicheng Xu$^{3}$ 
    \\
    \textbf{Chengtang Yao}$^{3}$ 
    \quad
    \textbf{Jianfeng Xiang}$^{1,3}$\footnotemark[2]
    \quad
    \textbf{Jiaolong Yang}$^{3}$\thanks{Corresponding author} 
    \\
	$^1${Tsinghua University}  \quad $^2${USTC} \quad $^3${Microsoft Research}
}

\usepackage[table]{xcolor}
\usepackage{colortbl}

\definecolor{bestcolor}{RGB}{206, 234, 195}
\definecolor{secondcolor}{RGB}{205, 226, 245}
\definecolor{thirdcolor}{RGB}{255, 244, 175}

\begin{document}
\maketitle
\vspace{-2em}
\begin{figure}[H]
  \centering
  \includegraphics[width=\linewidth]{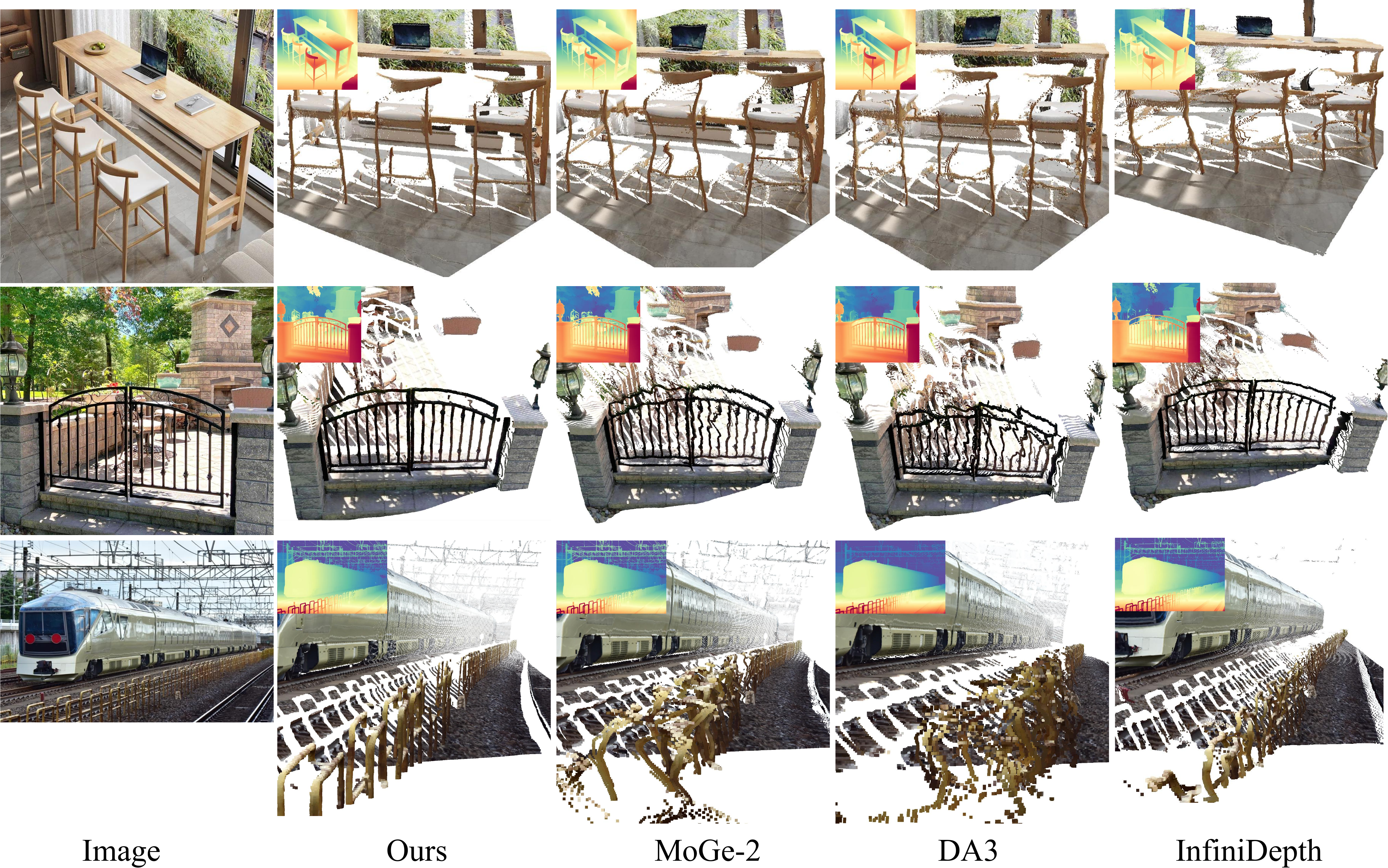}
  
  \caption{\small
  Our method produces geometrically undistorted and detail-preserving metric scale point maps from a single image. Compared to prior approaches~\cite{wang2025moge2accuratemonoculargeometry,lin2025depth3recoveringvisual,yu2026infinideptharbitraryresolutionfinegraineddepth}, it better preserves fine structures and recovers more accurate local geometry, leading to improved 3D fidelity in challenging cases.
  }
\label{fig:teaser}
\end{figure}
\begin{abstract}
Monocular geometry estimation has recently achieved impressive performance across diverse scenes. However, state-of-the-art models still face notable distortion in local 3D structure, especially in fine details, like thin structures and small objects. We attribute this limitation to an architectural mismatch: most current models decode 3D geometry within a 2D parameterization, where feature interactions are governed by image-plane proximity rather than true 3D spatial relationships. This inadvertently mixes features from geometrically distant surfaces, resulting in over-smoothed geometry particularly around thin or elongated structure. In this paper, we propose MoGe-3, a fine-detail monocular geometry estimation model with Self-Guided Sparse 3D Refinement (SSR) that lifts monocular geometry modeling from 2D image space to 3D space for high-fidelity metric-scale point maps. MoGe-3 lifts the coarse point map from a foundation base model onto a sparse voxel shell and refines it via SSR. The SSR employs sparse convolutions that aggregate features based on 3D spatial locality, avoiding feature mixing across depth discontinuities. Extensive experiments on diverse datasets demonstrate that MoGe-3 significantly outperforms existing approaches in recovering fine detailed 3D geometry across both quantitative metrics and qualitative visualizations. Project page: \url{https://qft-333.github.io/moge3page/}.
\end{abstract}
\section{Introduction}
Recovering the 3D geometry from a single RGB image has been a long-standing problem in computer vision, with broad applications in novel-view synthesis, AR/VR, and robotic manipulation. 
Building on advances in modern foundation backbones~\cite{oquab2024dinov2learningrobustvisual,siméoni2025dinov3}, recent monocular geometry estimation methods~\cite{DBLP:journals/corr/abs-1907-01341,visiontransformersfordenseprediction,bhat2023zoedepthzeroshottransfercombining,yin2023metric3dzeroshotmetric3d,Hu_2024,piccinelli2024unidepthuniversalmonocularmetric,yang2024depthanythingunleashingpower,yang2024depthv2,ke2024repurposingdiffusionbasedimagegenerators,bochkovskii2025depthprosharpmonocular,wang2025mogeunlockingaccuratemonocular,wang2025moge2accuratemonoculargeometry} have significantly improved zero-shot depth and point-map prediction. 
Despite this progress, we observe a systematic artifact: while current models produce plausible global layouts, they often struggle to faithfully reconstruct fine-scale geometry, particularly for thin or elongated structures, resulting in severe distortions such as twisted fences or smeared poles, as shown in Fig.~\ref{fig:teaser}.
Notably, such artifacts are often inconspicuous in 2D depth visualizations but become evident under viewpoint changes, exposing underlying 3D geometric distortions.

We hypothesize that this issue is due to a fundamental architectural mismatch: \textit{most current monocular reconstruction models decode 3D geometry within a 2D parameterization}. Feed-forward methods rely on 2D convolutional decoders~\cite{ DBLP:journals/corr/abs-1907-01341, visiontransformersfordenseprediction, bochkovskii2025depthprosharpmonocular,wang2025mogeunlockingaccuratemonocular, wang2025moge2accuratemonoculargeometry, Hu_2024, yang2024depthv2,yu2026infinideptharbitraryresolutionfinegraineddepth}, diffusion-based approaches employ VAE decoding~\cite{ke2024repurposingdiffusionbasedimagegenerators, fu2024geowizardunleashingdiffusionpriors}, and recent methods adopt MLP heads over patch latents or 2D UV queries~\cite{xu2025pixelperfectdepthsemanticsprompteddiffusion,yu2026infinideptharbitraryresolutionfinegraineddepth}. 
In all cases, feature interactions are governed by image-plane proximity rather than true 3D geometric relationships. This introduces a local smoothness prior that inadvertently mixes features from geometrically distant surfaces across depth discontinuities and over-smooths geometry, particularly around thin or elongated structures. Consequently, local structural integrity breaks down when 2D proximity diverges from true 3D relationships.

Our solution to this problem is to \textit{lift monocular geometry modeling from the 2D image plane to 3D space}. While retaining a strong 2D foundation model (e.g., DINOv2~\cite{oquab2024dinov2learningrobustvisual}) for appearance understanding, our method performs geometry modeling under a 3D inductive bias, where feature interactions are defined by spatial neighborhoods in 3D point space.
Through this design, image-adjacent pixels across depth discontinuities are decoupled, while image-disconnected pixels belonging to the same surface are associated via 3D proximity. This new paradigm addresses the fundamental causes of local structure distortions in current monocular geometry estimation models.

MoGe-3 consists of two main components: a foundational geometry estimator as the base model and a \textbf{S}elf-guided \textbf{S}parse 3D \textbf{R}efiner (SSR). 
Since a 3D point map occupies a thin manifold within a sparse volume, traditional implementation of a 3D-aware decoder at full image resolution is computationally inefficient. We therefore operate directly on the occupied scene manifold using 3D sparse convolutions. 
Specifically, we initialize the base geometry estimator from MoGe-2~\cite{wang2025moge2accuratemonoculargeometry} to generate an initial point map. We then perform iterative refinement using SSR in a self-guided manner, where the sparse 3D U-Net structure at each step is determined by the previous prediction.
Before being processed by the U-Net, the point map is voxelized in a log-depth parameterized coordinate space for scale-consistent discretization. 2D features from the geometry estimator are also injected into the 3D U-Net bottlenecks. 
This sparse formulation operates only on occupied regions, making computation linear in point occupancy, and enables 3D-aware refinement without sacrificing the capacity of the 2D backbone.

We train our model on a large mixture of pixel-accurate synthetic data and diverse real-world datasets. Across standard zero-shot benchmarks, our default ViT-L model achieves state-of-the-art performance, consistently outperforming existing approaches in both local and global geometric accuracy. We further train a scaled ViT-G variant to show that SSR remains effective on top of a stronger foundation backbone. Beyond quantitative improvements, MoGe-3 produces clearly superior qualitative results, with more accurate fine-scale geometry, cleaner and more consistent geometric structures, and significantly reduced distortions in regions of complex geometry like thin and elongated structures, where prior methods often fail.
This bridges the gap between visually plausible depth and faithful 3D reconstruction.
\section{Related Works}
\label{sec:related}
\paragraph{Monocular depth \& geometry estimation.}
Monocular geometry estimation aims to recover depth maps or point clouds from a single image. Early works such as MiDaS~\cite{DBLP:journals/corr/abs-1907-01341} and DPT~\cite{visiontransformersfordenseprediction} established large-scale training and transformer-based decoding for relative depth estimation. Subsequent approaches extended this paradigm toward metric and generalizable geometry estimation, including ZoeDepth~\cite{bhat2023zoedepthzeroshottransfercombining}, Metric3D / Metric3D-v2~\cite{yin2023metric3dzeroshotmetric3d,Hu_2024}, Depth Anything / V2~\cite{yang2024depthanythingunleashingpower,yang2024depthv2}, and UniDepth~\cite{piccinelli2024unidepthuniversalmonocularmetric}. The MoGe family~\cite{wang2025mogeunlockingaccuratemonocular,wang2025moge2accuratemonoculargeometry} further reformulates the task as affine-invariant point-map regression.

Despite their strong performance, these approaches still suffer from noticeable geometric fidelity distortions. To address this problem, Depth Pro~\cite{bochkovskii2025depthprosharpmonocular} turns to high-resolution image input and multi-scale vision encoder architecture. Marigold~\cite{ke2024repurposingdiffusionbasedimagegenerators} and GeoWizard~\cite{fu2024geowizardunleashingdiffusionpriors} introduce powerful generative priors for monocular geometry through VAE-based decoding. Pixel-Perfect Depth~\cite{xu2025pixelperfectdepthsemanticsprompteddiffusion} further improves edge fidelity by directly performing diffusion generation in the pixel space, avoiding VAE-compression-induced artifacts. InfiniDepth~\cite{yu2026infinideptharbitraryresolutionfinegraineddepth}, instead, represents depth as a neural implicit field and adopts 2D UV queries for finer-grained depth estimation.

Although these methods significantly improve reconstruction quality, they still rely on decoding 3D geometry within a 2D parameterization. As a result, feature interactions are primarily determined by image-plane proximity rather than actual 3D spatial relationships. This mismatch can limit geometric accuracy when neighboring pixels in the image are distant in 3D space, or vice versa.

\vspace{-6pt}
\paragraph{Sparse 3D Aggregation.} Sparse 3D aggregation focuses on explicit 3D structure and aggregates features according to 3D spatial locality. Prior methods mainly use sparse 3D convolutional networks to operate directly in voxel space while avoiding empty-space overhead, achieving strong performance in different downstream tasks, such as scene understanding~\cite{graham20173dsemanticsegmentationsubmanifold,choy20194dspatiotemporalconvnetsminkowski,Yan2018SECONDSE,tang2020searchingefficient3darchitectures} and 3D generation~\cite{xiang2025native}. Some methods also use sparse 3D aggregation to refine the noisy point cloud inputs~\cite{wang2024sparse,rich20213dvnet}, significantly improving reconstruction fidelity and overall quality. To the best of our knowledge, we are the first to lift monocular geometry modeling from the 2D image plane to 3D space and aggregate features via sparse 3D convolutions, mitigating erroneous feature mixing across depth discontinuities.
\section{Method}
\label{headings}
As illustrated in Figure~\ref{fig:overview}, MoGe-3 consists of two main components: a base model initialized from MoGe-2~\cite{wang2025moge2accuratemonoculargeometry} that produces an initial point map prediction, and a Self-Guided Sparse 3D Refiner (SSR) that further enhances geometric fidelity in 3D space.
The base model also extracts a dense feature map $\boldsymbol{F}$ from a DINOv2 encoder~\cite{oquab2024dinov2learningrobustvisual}, which we reuse in our refinement stage. 

\subsection{Base Model}
\label{sec:preliminaries}
Given an input image, our base model extracts a dense feature map $\boldsymbol{F}$ and produces a dense per-pixel 3D point map $\boldsymbol{P}$ together with auxiliary predictions including a validity mask, surface normals, and a global metric scale, as shown in Figure~\ref{fig:overview}(a). Same as \cite{wang2025moge2accuratemonoculargeometry}, $\boldsymbol{P}$ is supervised only up to an unknown
\emph{affine transform along the optical axis}, i.e.\ the loss is
invariant under
\begin{equation}
(X, Y, Z) \;\mapsto\; \big(s X,\; s Y,\; s Z + t\big),
\qquad s > 0,\; t \in \mathbb{R}.
\end{equation}
The \emph{metric point map} is constructed by multiplying the predicted metric scale with $\boldsymbol{P}$.

\subsection{SSR: Self-Guided Sparse 3D Refiner}
\label{sec:method}
The SSR module takes the initial point map $\boldsymbol{P}$ as input and iteratively refines its geometry. To normalize the point cloud with different range of $(X_{ij}, Y_{ij})$ into a uniform space and focus on depth refinement, we employ a factorized representation $\boldsymbol{q}_{ij}$ for each point $\boldsymbol{p}_{ij}$ to :
\begin{equation}
\boldsymbol{q}_{ij} = (u_{ij} , v_{ij}, \zeta_{ij}) = \left(\frac{X_{ij}}{Z_{ij}}, \frac{Y_{ij}}{Z_{ij}}, \log Z_{ij}\right),
\label{eq:factorisation}
\end{equation}
where $i$ and $j$ are the row and column indices of the corresponding pixel. In this formulation, the image-axis coordinates $(u, v)$ remain fixed throughout the process, allowing the refiner to concentrate on predicting residuals for the log-depth $\zeta$, starting from the initial $\zeta^{(0)}$.

\begin{figure}[t]
    \centering
    \includegraphics[width=\linewidth]{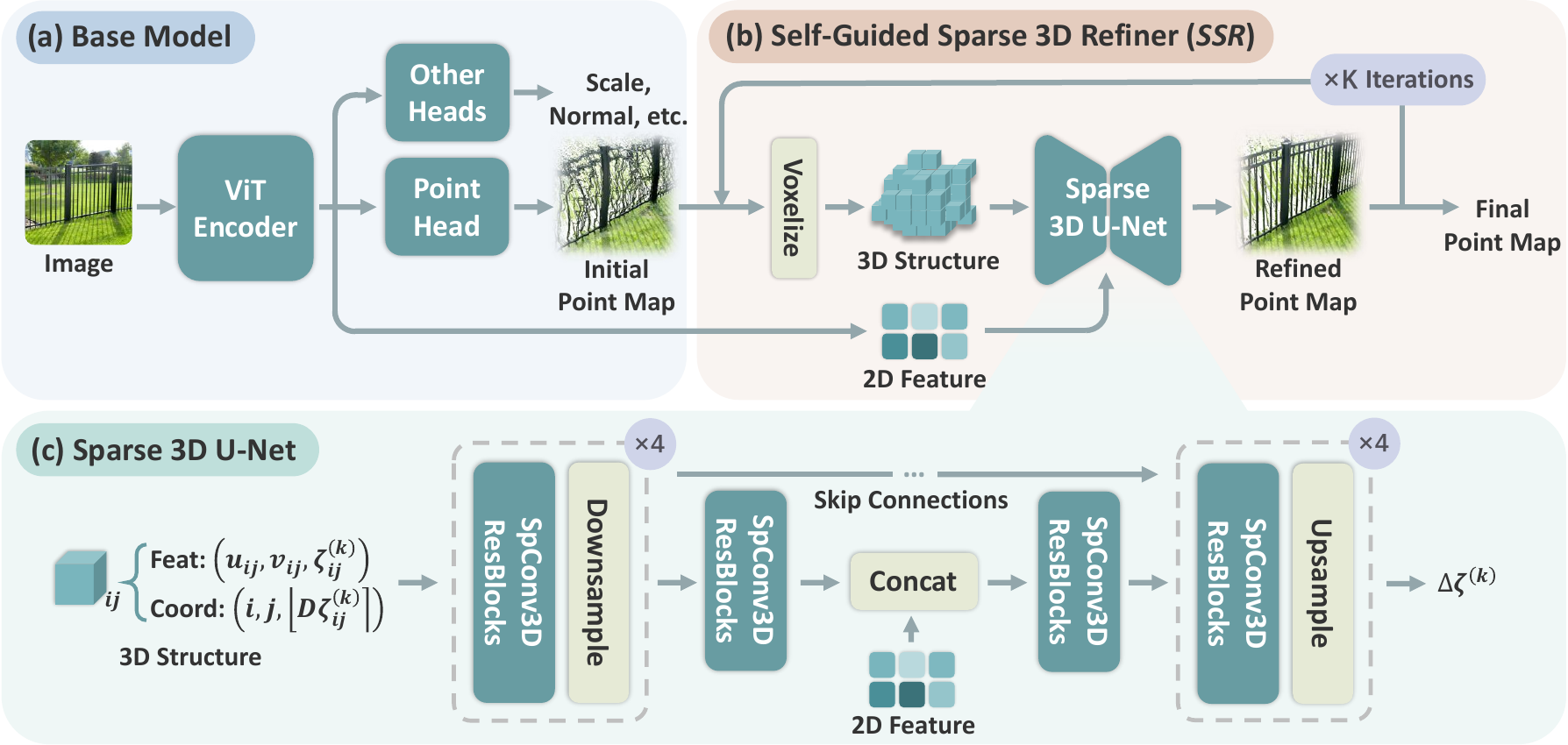}
    \caption{\textbf{Method overview.} (a) The \textbf{Base model} provides an initial point map and 2D features. (b) The \textbf{ SSR module} iteratively refines this geometry: current estimates are discretized into sparse voxel shells that guide 3D inference, which then updates the point map for the subsequent cycle. (c) The \textbf{Sparse 3D U-Net} employs multi-scale sparse 3D convolutions, injecting 2D features from base model, and predicts log-depth residuals.}
    \label{fig:overview}
\end{figure}

As illustrated in Figure~\ref{fig:overview}(b), each iteration consists of three steps: (1) Self-guided voxelization~\ref{sec:self-guided-voxelization}, where the current point map is discretized into a sparse 3D voxel shell; (2) Sparse 3D U-Net~\ref{sec:sparse-unet}, which performs multi-scale feature extraction and 2D-3D feature integration; and (3) Iterative update~\ref{sec:iterative}, which applies the predicted residuals to refine the geometry.

\subsubsection{Self-Guided Voxelization}
\label{sec:self-guided-voxelization}
To lift monocular geometry modeling from the 2D image plane to 3D space, we first transform the point map into a sparse voxel set, where each pixel is corresponding to exactly one voxel. In this construction, the image axes $(i, j)$ provide a natural discretization, while the depth axis is quantized by scaling the log-depth $\zeta_{ij}^{(k)}=\log Z_{ij}^{(k)}$ with a specific resolution $D$. The sparse voxel set at iteration $k$ is defined as:
\begin{equation}
\mathcal{V}^{(k)} \;=\; \left\{\, \boldsymbol{c}_{ij}^{(k)}\right\}, 
\qquad
\boldsymbol{c}_{ij}^{(k)}= \left(i,\, j,\, \left\lfloor D \cdot \zeta^{(k)}_{ij} \right\rceil\right)
\end{equation}
where $\lfloor \cdot \rceil$ denotes rounding to the nearest integer. 
Each occupied voxel is associated with a factorized point representation $\boldsymbol{q}^{(k)}_{ij}$, which serves as a 3-channel feature input.

This construction yields two key properties:
(1) \textit{Projection-aligned \& scale-invariant shell:} $\mathcal{V}^{(k)}$ is coherent with the image resolution and depth relative precision, forming a thin manifold with exactly $HW$ occupied voxels. The shell density and connectivity remain invariant to metric scale transformations, ensuring consistent geometric representation across diverse scene scales.
(2) \textit{Geometry-coherent receptive field:} Two pixels (voxels) are 3D-adjacent in $\mathcal{V}^{(k)}$ if and only if they are image-adjacent and their log-depths satisfy $|\zeta_{ij}^{(k)} - \zeta_{i'j'}^{(k)}| \le 1/D$. This naturally decouples features across depth discontinuities, as pixels from different surfaces land in separate voxels, preventing the "bleeding" and "twisted" artifacts inherent in 2D-based feature mixing.

\subsubsection{Sparse 3D U-Net}
\label{sec:sparse-unet}
We build a sparse 3D U-Net to process the lifted monocular geometry in 3D space. As illustrated in Figure~\ref{fig:overview}(c), our sparse 3D U-Net first encodes the sparse voxel set $\mathcal{V}^{(k)}$ with multiple sparse 3D convolution blocks, each followed by a downsampling operation. Then, at the lowest resolution level, the encoded voxels are concatenated with 2D  features $\boldsymbol{F}$ from base model. The concatenated results are then used to decode a \textit{scalar log-depth residual} $\Delta \zeta_{ij}^{(k)}$ for point map refinement.

\vspace{-6pt}
\paragraph{Dynamic-sparsity downsample \& upsample.} 
Different from the vanilla 2D U-Net, which performs downsampling and upsampling uniformly in image space, our method enables geometry-dependent sparse structures across resolution levels through sparse voxel representations. At each level $l$, the sparse voxel set is defined as $\mathcal{V}^{(k),l} = \left\{ \left\lfloor \tfrac{1}{2^l} \boldsymbol{c}_{ij}  \right\rfloor \right\}$. While the input level ($l=0$) contains exactly $HW$ voxels, the number and distribution of voxels at coarser levels ($l \ge 1$) vary with the underlying scene geometry. In particular, structures that collapse to the same image-space location can still remain separated along the depth dimension, leading to multiple occupied voxels around depth discontinuities and naturally reducing foreground-background feature mixing during multi-scale processing.

\vspace{-6pt}
\paragraph{2D feature injection.} Since voxel features only encode geometric coordinates and lack high-level scene understanding, we further inject the features $\boldsymbol{F}$ from base model into the U-Net bottleneck for efficient fusion at reduced spatial resolution. At this level, 2D tokens are mapped to bottleneck voxels based on their image-axis coordinates. Voxels sharing an image location but differing in depth duplicate the same 2D feature, allowing 3D convolutions to read from rich appearance priors.
\vspace{-6pt}
\paragraph{Output layer initialization.} The U-Net predicts a \textit{scalar log-depth residual} $\Delta \zeta_{ij}^{(k)}$ corresponding to each input voxel. In order to stabilize the optimization at the start of training, we zero-initialize the output layer of sparse 3D U-Net, allowing identity mapping.

\subsubsection{Iterative update}
\label{sec:iterative}
The predicted log-depth residuals $\Delta \zeta^{( k)}$ are added to the current estimate to update the factorized representation:
\begin{equation}
\boldsymbol{q}^{(k+1)}_{ij} = \left( u_{ij}, v_{ij}, \zeta_{ij}^{(k)} + \Delta\zeta_{ij}^{(k)} \right).
\end{equation}
The updated $\boldsymbol{q}^{(k+1)}$ defines the voxel shell for the subsequent iteration $k+1$. This self-guiding loop progressively refines the spatial locality of the 3D convolutions, allowing the refiner to operate on increasingly precise surface boundaries. 

After $K$ iterations, the final point map $\boldsymbol{p}^{(K)}$ is recovered in Euclidean space via:
\begin{equation}
\boldsymbol{p}_{ij}^{(K)}
= e^{\zeta_{ij}^{(K)}}\cdot \left(u_{ij},v_{ij},1\right).
\label{eq:factorized_to_euclidean}
\end{equation}
\vspace{-10pt}
\subsection{Losses and Training}
\label{sec:losses}

We optimize a composite objective that generally follows the formulation in \cite{wang2025moge2accuratemonoculargeometry}. For each SSR iteration $k \in \{0, \dots, K\}$, we define a geometric objective $\mathcal{L}_{\mathrm{geo}}^{(k)}$ that combines a global affine-invariant loss $\mathcal{L}_{\mathrm{global}}^{(k)}$, a multi-scale local loss $\mathcal{L}_{\mathrm{local}}^{(k)}$, and an edge-angle loss $\mathcal{L}_{\mathrm{edge}}^{(k)}$.
In addition to these per-iteration terms, we supervise the auxiliary outputs inherited from the base model, including mask, surface normal, and global scale predictions using their respective loss terms $\mathcal{L}_{\mathrm{mask}}$, $\mathcal{L}_{\mathrm{nml}}$, and $\mathcal{L}_{\mathrm{scale}}$. Their detailed descriptions can be found in Appendix~\ref{sec:app-losses}. The full loss objective is defined as:
\begin{equation}
\mathcal{L} = \sum_{k=0}^{K} \mathcal{L}_{\mathrm{geo}}^{(k)} + \lambda_{\mathrm{m}} \mathcal{L}_{\mathrm{mask}} + \lambda_{\mathrm{n}} \mathcal{L}_{\mathrm{nml}} + \lambda_{\mathrm{s}} \mathcal{L}_{\mathrm{scale}}.
\label{eq:total-loss}
\end{equation}

The model is trained in two stages. \textbf{Stage 1} (\emph{warm-up}) detaches gradients from the base model features to the refiner. Combined with the zero-initialization of the output layer, this stage encourages the refiner to strictly learn residual corrections starting from a known-good initialization. \textbf{Stage 2} (\emph{joint fine-tuning}) enables end-to-end training of the entire pipeline. We empirically find this two-stage training strategy essential, as it prevents large early-stage gradients of refiner from corrupting the representation of well-pretrained base model. We further partition the training data across components: the refiner receives gradients only from synthetic samples, whose pixel-accurate ground truth is essential for learning sharp and geometrically correct residual corrections, while the base model is supervised on the full mixture of synthetic and real-world data to retain in-the-wild generalization.
Detailed hyper-parameters are provided in Appendix~\ref{sec:app-training}.
\section{Experiments}
\label{sec:experiments}

\subsection{Implementation Details}
We train our model on a large-scale mixture of synthetic and real-world datasets; see Appendix~\ref{sec:app-training} for the training data details. The model is trained using AdamW with a staged training schedule.
The full training details, including loss weights, learning rates, and iteration settings, are provided in Appendix~\ref{sec:app-training}. \emph{\textbf{All our code, model, training and evaluation scripts will be made publicly available.}}

Unless otherwise specified, our default model is the full two-stage-trained model with a DINOv2 ViT-L backbone and $K=3$ refinement iterations at inference. We also train a scaled-up variant with a DINOv2 ViT-G backbone.

\subsection{Evaluation Setting}
\label{sec:exp-setup}

\paragraph{Evaluation datasets.}
We report zero-shot evaluation results on 9 benchmarks that span synthetic and real, indoor and outdoor, dense and sparse-LiDAR ground truth:
Synth4K~\cite{yu2026infinideptharbitraryresolutionfinegraineddepth}, Spring~\cite{Mehl2023Spring},
NYUv2~\cite{Silberman2012nyuv2},
KITTI~\cite{Uhrig2017kitti}, 
ETH3D~\cite{Schops2019ETH3D}, 
DIODE~\cite{diode_dataset},
Sintel~\cite{Butler2012sintel}, 
iBims-1~\cite{ibim1_1}, and HAMMER~\cite{jung2023hammer}.

\vspace{-6pt}
\paragraph{Evaluation metrics.}
We report three types of evaluation metrics:
(i) Global metrics evaluate full-image geometry under three
settings: affine-invariant point maps (\emph{Relative Point}),
affine-invariant depth (\emph{Relative Depth}), and metric depth
(\emph{Metric Depth}). For each setting, we report relative error (Rel) and threshold accuracy $\delta$, using thresholds of $0.25$ for points and $1.25$ for depth.
(ii) Local (fine-detail) metrics measure accuracy in
regions with thin structures and complex geometry. We
automatically extract region-level masks that select
small segments containing fine-detail geometry structures (see Appendix~\ref{sec:appendix_fine_grained_mask} for details). For each region, we perform per-segment
alignment and compute Rel
and $\delta$ metrics, using stricter thresholds of $0.01$ for point maps and $1.01$ for
depth maps. The reported numbers are the average over all selected regions.
(iii) Boundary F1 (radius
$r{=}1$) evaluates depth discontinuity sharpness by comparing predicted and ground-truth boundary maps.

\vspace{-6pt}
\paragraph{Baselines.}
We compare against seven recent state-of-the-art methods:
Depth Pro~\cite{bochkovskii2025depthprosharpmonocular},
UniDepth V2~\cite{piccinelli2025unidepthv2},
Depth Anything 3~\cite{lin2025depth3recoveringvisual},
UniK3D~\cite{piccinelli2025unik3duniversalcameramonocular},
Pixel-Perfect Depth~\cite{xu2025pixelperfectdepthsemanticsprompteddiffusion},
InfiniDepth~\cite{yu2026infinideptharbitraryresolutionfinegraineddepth},
and MoGe-2~\cite{wang2025moge2accuratemonoculargeometry}. 
Pixel-Perfect Depth and InfiniDepth predict relative depth and do not estimate camera intrinsics. We therefore omit metric-depth evaluation for these methods and reconstruct 3D point maps using camera intrinsics estimated by MoGe-2 on the same image for point-based metrics. All other baselines use their own predicted intrinsics.

\begin{table*}[t!]
    \scriptsize
    \centering
    \caption{Zero-shot evaluation and comparison with state-of-the-art methods. Metrics are averaged over multiple benchmarks (see text for details). Results are visualized using a color gradient from \colorbox[rgb]{0.780,1.000,0.780}{green} (best) to \colorbox[rgb]{1.000,0.780,0.780}{red} (worst); \colorbox[rgb]{0.850,0.850,0.850}{\textcolor{gray}{-}} indicates non-applicable. Unless otherwise specified, our method uses the default ViT-L backbone; the ViT-G rows report our scaled-up SSR model.See Appendix~\ref{sec:appendix_more_results} for per-dataset results.} 
    \begin{tabular}{l|cc|cc|c|cc|cc|cc}
    \specialrule{.12em}{0em}{0em}

    \multirow{3}{*}{\textbf{Method}}
    & \multicolumn{4}{c|}{Local}
    & \multicolumn{1}{c|}{\!\!\!Boundary\!\!\!}
    & \multicolumn{6}{c}{Global} \\

    \cline{2-5} \cline{6-6} \cline{7-12}

    & \multicolumn{2}{c|}{\scriptsize Depth}
    & \multicolumn{2}{c|}{\scriptsize Point}
    & \multirow{2}{*}{\scriptsize F1$\uparrow$}
    & \multicolumn{2}{c|}{\scriptsize Relative Depth}
    & \multicolumn{2}{c|}{\scriptsize Relative Point}
    & \multicolumn{2}{c}{\scriptsize Metric Depth} \\

    \cline{2-5} \cline{7-12}

    & \scriptsize Rel\textsuperscript{d}$\downarrow$
    & \scriptsize $\delta_{0.01}^\text{d}\uparrow$
    & \scriptsize Rel\textsuperscript{p}$\downarrow$
    & \scriptsize $\delta_{0.01}^\text{p}\uparrow$
    &
    & \scriptsize Rel\textsuperscript{d}$\downarrow$
    & \scriptsize $\delta_{1}^\text{d}\uparrow$
    & \scriptsize Rel\textsuperscript{p}$\downarrow$
    & \scriptsize $\delta_{1}^\text{p}\uparrow$
    & \scriptsize Rel\textsuperscript{d}$\downarrow$
    & \scriptsize $\delta_{1}^\text{d}\uparrow$ \\
    \hline
        Depth Pro & \cellcolor[rgb]{0.906,1.000,0.780} 4.15 & \cellcolor[rgb]{0.906,1.000,0.780} 46.9 & \cellcolor[rgb]{0.843,1.000,0.780} 3.01 & \cellcolor[rgb]{0.906,1.000,0.780} 45.2 & \cellcolor[rgb]{0.843,1.000,0.780} 16.3 & \cellcolor[rgb]{1.000,0.843,0.780} 8.10 & \cellcolor[rgb]{1.000,0.853,0.780} 91.3 & \cellcolor[rgb]{1.000,0.906,0.780} 11.1 & \cellcolor[rgb]{1.000,0.927,0.780} 87.6 & \cellcolor[rgb]{1.000,0.780,0.780} 29.0 & \cellcolor[rgb]{1.000,0.780,0.780} 54.0 \\
        UniDepth V2 & \cellcolor[rgb]{1.000,0.906,0.780} 5.60 & \cellcolor[rgb]{1.000,0.969,0.780} 45.3 & \cellcolor[rgb]{1.000,0.843,0.780} 4.60 & \cellcolor[rgb]{1.000,0.843,0.780} 42.6 & \cellcolor[rgb]{1.000,0.906,0.780} 14.5 & \cellcolor[rgb]{0.969,1.000,0.780} 6.98 & \cellcolor[rgb]{0.927,1.000,0.780} 92.8 & \cellcolor[rgb]{0.969,1.000,0.780} 10.2 & \cellcolor[rgb]{1.000,1.000,0.780} 88.6 & \cellcolor[rgb]{1.000,0.868,0.780} 23.7 & \cellcolor[rgb]{1.000,0.956,0.780} 71.2 \\
        Depth Anything 3 & \cellcolor[rgb]{1.000,0.780,0.780} 10.5 & \cellcolor[rgb]{1.000,0.843,0.780} 41.8 & \cellcolor[rgb]{1.000,0.780,0.780} 7.39 & \cellcolor[rgb]{1.000,0.969,0.780} 44.7 & \cellcolor[rgb]{1.000,0.780,0.780} 6.61 & \cellcolor[rgb]{1.000,0.906,0.780} 7.89 & \cellcolor[rgb]{1.000,0.927,0.780} 91.4 & \cellcolor[rgb]{1.000,0.969,0.780} 10.6 & \cellcolor[rgb]{1.000,0.853,0.780} 87.5 & \cellcolor[rgb]{0.956,1.000,0.780} 17.4 & \cellcolor[rgb]{1.000,0.868,0.780} 70.1 \\
        UniK3D & \cellcolor[rgb]{1.000,0.969,0.780} 5.52 & \cellcolor[rgb]{0.969,1.000,0.780} 46.2 & \cellcolor[rgb]{1.000,0.969,0.780} 4.37 & \cellcolor[rgb]{0.969,1.000,0.780} 45.0 & \cellcolor[rgb]{1.000,0.969,0.780} 15.1 & \cellcolor[rgb]{0.906,1.000,0.780} 6.95 & \cellcolor[rgb]{0.927,1.000,0.780} 92.8 & \cellcolor[rgb]{0.906,1.000,0.780} 9.63 & \cellcolor[rgb]{0.927,1.000,0.780} 89.6 & \cellcolor[rgb]{1.000,0.956,0.780} 17.9 & \cellcolor[rgb]{0.868,1.000,0.780} 77.8 \\
        Pixel-Perfect Depth & \cellcolor[rgb]{0.969,1.000,0.780} 5.24 & \cellcolor[rgb]{1.000,0.780,0.780} 39.8 & \cellcolor[rgb]{1.000,0.906,0.780} 4.39 & \cellcolor[rgb]{1.000,0.780,0.780} 39.1 & \cellcolor[rgb]{1.000,0.843,0.780} 12.7 & \cellcolor[rgb]{1.000,0.969,0.780} 7.17 & \cellcolor[rgb]{1.000,1.000,0.780} 92.5 & \cellcolor[rgb]{1.000,0.780,0.780} 20.3 & \cellcolor[rgb]{1.000,0.780,0.780} 69.8 & \cellcolor[rgb]{0.850,0.850,0.850} \textcolor{gray}{-} & \cellcolor[rgb]{0.850,0.850,0.850} \textcolor{gray}{-} \\
        InfiniDepth & \cellcolor[rgb]{1.000,0.843,0.780} 5.82 & \cellcolor[rgb]{1.000,0.906,0.780} 43.2 & \cellcolor[rgb]{0.969,1.000,0.780} 4.29 & \cellcolor[rgb]{1.000,0.906,0.780} 43.7 & \cellcolor[rgb]{0.780,1.000,0.780} 19.3 & \cellcolor[rgb]{1.000,0.780,0.780} 8.15 & \cellcolor[rgb]{1.000,0.780,0.780} 91.1 & \cellcolor[rgb]{1.000,0.843,0.780} 11.6 & \cellcolor[rgb]{1.000,0.927,0.780} 87.6 & \cellcolor[rgb]{0.850,0.850,0.850} \textcolor{gray}{-} & \cellcolor[rgb]{0.850,0.850,0.850} \textcolor{gray}{-} \\
        MoGe-2 & \cellcolor[rgb]{0.843,1.000,0.780} 3.81 & \cellcolor[rgb]{0.843,1.000,0.780} 47.1 & \cellcolor[rgb]{0.906,1.000,0.780} 3.19 & \cellcolor[rgb]{0.843,1.000,0.780} 46.6 & \cellcolor[rgb]{0.969,1.000,0.780} 15.6 & \cellcolor[rgb]{0.843,1.000,0.780} 5.98 & \cellcolor[rgb]{0.853,1.000,0.780} 94.1 & \cellcolor[rgb]{0.843,1.000,0.780} 8.73 & \cellcolor[rgb]{0.853,1.000,0.780} 90.4 & \cellcolor[rgb]{0.868,1.000,0.780} 15.6 & \cellcolor[rgb]{0.956,1.000,0.780} 77.3 \\
        ${Ours}$ (ViT-L)& \cellcolor[rgb]{0.780,1.000,0.780} 3.76 & \cellcolor[rgb]{0.780,1.000,0.780} 52.5 & \cellcolor[rgb]{0.780,1.000,0.780} 2.79 & \cellcolor[rgb]{0.780,1.000,0.780} 55.9 & \cellcolor[rgb]{0.906,1.000,0.780} 16.0 & \cellcolor[rgb]{0.780,1.000,0.780} 5.52 & \cellcolor[rgb]{0.780,1.000,0.780} 94.7 & \cellcolor[rgb]{0.780,1.000,0.780} 7.91 & \cellcolor[rgb]{0.780,1.000,0.780} 92.0 & \cellcolor[rgb]{0.780,1.000,0.780} 15.0 & \cellcolor[rgb]{0.780,1.000,0.780} 82.7 \\

        \hline
        ${Ours}$ (ViT-G) & 3.66 & 54.1 
        & 2.80 & 56.8
        & 17.3
        & 4.80 & 95.6 
        & 7.43 & 92.4 
        & 15.8 & 82.7 \\
        \specialrule{.12em}{0em}{0em}
    \end{tabular}
    \vspace{-10pt}
    \label{tab:main}
\end{table*}
\subsection{Quantitative Evaluation}
\label{sec:exp-zeroshot}
Table~\ref{tab:main} reports zero-shot evaluation results averaged across the evaluation benchmarks. Global metrics are averaged over all nine test datasets. Local metrics are averaged over Spring and Synth4K, as region-level mask extraction requires dense ground truth with fine-detail structures. Boundary F1 is averaged over Spring, iBims-1, Sintel, and HAMMER. These four datasets are representative in that they include both synthetic (Spring, Sintel) and real-world (iBims-1, HAMMER) sources, and contain abundant high-fidelity depth edges for boundary evaluation.

\paragraph{Global metrics.}
Our method achieves the best global performance across all three evaluation settings, outperforming the previous best method MoGe-2~\cite{wang2025moge2accuratemonoculargeometry} by a clear margin. This demonstrates that joint training with the SSR module not only enhances local details but also improves overall geometric accuracy.

\vspace{-6pt}
\paragraph{Local (fine-detail) metrics.}
The advantage of our method is most pronounced on local fine-detail metrics, where we \emph{substantially outperform all baselines under the strict threshold accuracy $\delta_{0.01}$}. This confirms that the 3D sparse refinement effectively resolves thin structures and complex geometry that 2D approaches struggle with.

\vspace{-6pt}
\paragraph{Boundary F1.}
InfiniDepth~\cite{yu2026infinideptharbitraryresolutionfinegraineddepth} achieves the best boundary sharpness by leveraging explicit 2$\times$ higher-resolution query processing to enhance image-space edge fidelity. Our method ranks third overall, and is comparable to Depth~Pro~\cite{bochkovskii2025depthprosharpmonocular} which operates at $1536^2$ native resolution, nearly $3.4\times$ higher than our $840^2$.

\begin{figure*}
    \centering
    \includegraphics[width=\linewidth]{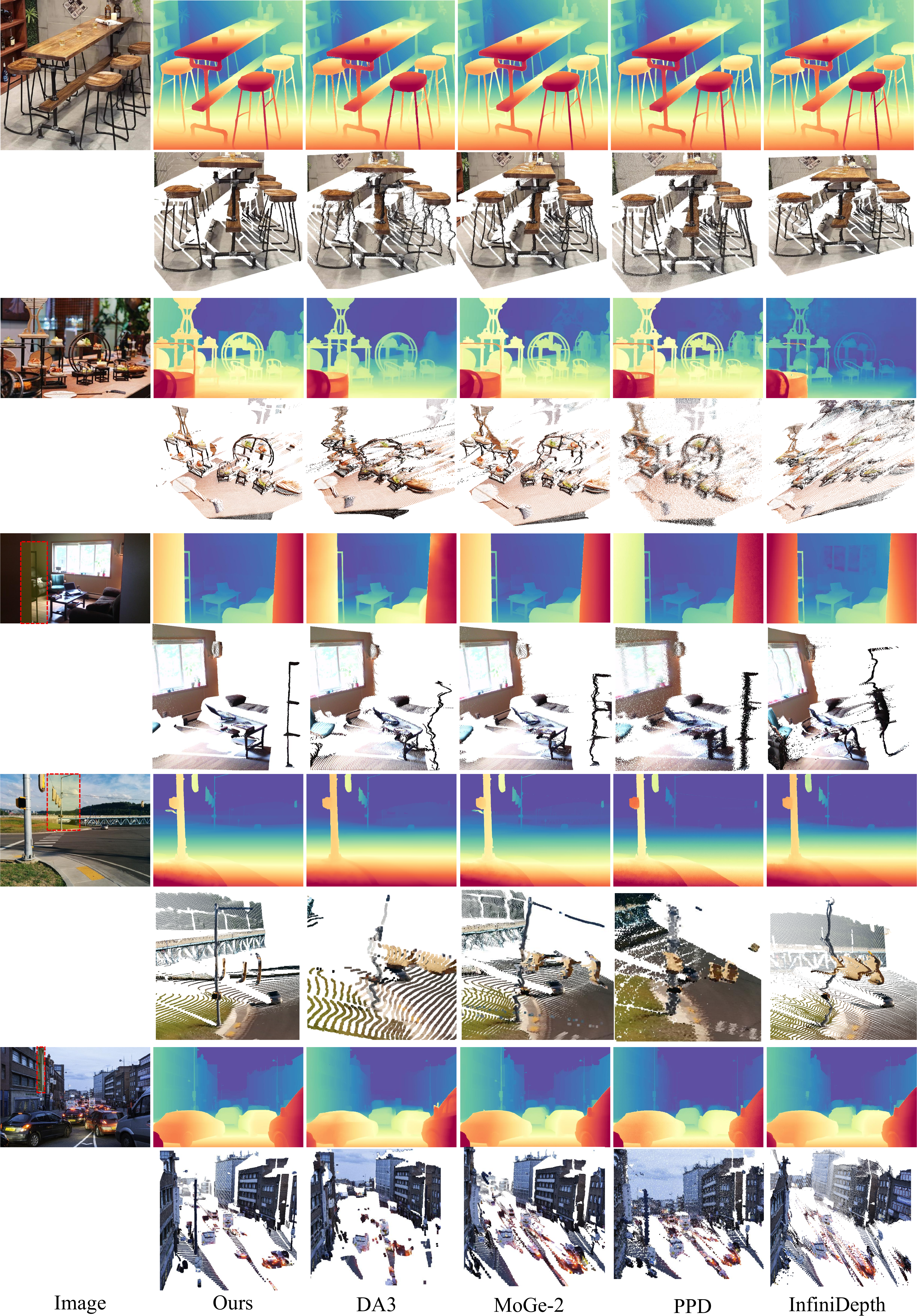}
    \caption{Qualitative comparison of point maps and disparity maps on unseen test images. Since Pixel-Perfect Depth  (PPD)~\cite{xu2025pixelperfectdepthsemanticsprompteddiffusion} and InfiniDepth~\cite{yu2026infinideptharbitraryresolutionfinegraineddepth}
 do not predict camera intrinsics, we use the intrinsics predicted by MoGe-2~\cite{wang2025moge2accuratemonoculargeometry}
 to lift their depth predictions into 3D. Our method produces noticeably higher-fidelity 3D geometry with reduced distortions, particularly in complex regions and thin structures. \emph{More visual comparisons can be found in Appendix~\ref{sec:appendix_visual}}.}
    \label{fig:qualitative_comparison}
\end{figure*}

It should be noted that boundary sharpness is a 2D, pixel-level metric that mainly reflects image-space depth edge fidelity and \emph{does not fully capture 3D geometric quality}. As shown in Fig.~\ref{fig:qualitative_comparison}, our method produces 3D point maps with much better structural fidelity. This is because the 3D refiner sharpens depth discontinuities by resolving local geometry in 3D space, where occluding and occluded surfaces are explicitly separated into distinct voxels rather than being blended across adjacent pixels.

\vspace{-6pt}
\paragraph{Scaling to a larger backbone.}
To examine whether SSR remains effective with stronger 2D foundation models, we train a scaled-up variant where SSR is applied on top of a MoGe-2-style base model with a DINOv2 ViT-G backbone. As shown in Table~\ref{tab:main}, the ViT-G model further improves over our ViT-L setting across most metric categories, indicating that SSR is complementary to backbone scaling and can effectively leverage the richer representations provided by larger encoders.

\begin{wraptable}{r}{0.5\linewidth}
\vspace{-14pt}
\centering
\scriptsize
\caption{Runtime comparison on a single A100 GPU with FP16.}
\vspace{-4pt}
\begin{tabular}{lcc}
\toprule
Method & Resolution & Latency (ms) \\
\midrule
MoGe-2 (ViT-L) & $700^2$ & 39 \\
Depth Anything 3 & $630{\times}840$ & 213 \\
Depth Pro & $1536^2$ & 142 \\
InfiniDepth & $512{\times}672$ & 133 \\
Pixel-Perfect Depth & $512{\times}672$ & 275 \\
\midrule
Ours (ViT-L, 3-step) & $700^2$ & 121 \\
Ours (ViT-G, 3-step) & $700^2$ & 177 \\
\bottomrule
\end{tabular}
\label{tab:runtime}
\vspace{-10pt}
\end{wraptable}

\vspace{-6pt}
\paragraph{Runtime analysis.}
Table~\ref{tab:runtime} reports inference latency. With $K{=}0$, our model has identical latency to MoGe-2 since they share the same base architecture. The 3-step SSR adds moderate overhead, bringing the ViT-L model to 121\,ms per frame, which is still faster than prior detail-enhancement methods while achieving better accuracy across most geometric metrics. The scaled-up ViT-G variant runs at 177\,ms, still competitive in speed and further improving quality. Detailed results are provided in Appendix~\ref{sec:appendix_more_results}.

\subsection{Qualitative Evaluation}
\label{sec:visual}

Fig.~\ref{fig:qualitative_comparison} presents qualitative comparisons of point maps and disparity maps against previous methods. For methods that do not predict camera intrinsics, we use the intrinsics predicted by MoGe-2~\cite{wang2025moge2accuratemonoculargeometry} to lift their depth predictions into 3D point maps. 
Our method produces sharp disparity boundaries in image space while maintaining faithful 3D geometry, with accurate reconstruction of complex regions with thin structures. In contrast, although prior methods often appear sharp in 2D depth maps, their reconstructed 3D point maps exhibit noticeable structural distortions. To the best of our knowledge, we are the first monocular geometry estimation method that produces visually plausible depth while also maintaining high-fidelity 3D geometry.

\vspace{-6pt}
\subsection{Ablation studies}
\label{sec:exp-ablation}
For the ablation studies on voxel resolution $D$, visual feature injection, and the 2D convolution variant, we use a refiner-only setting where the base model is frozen and only the SSR module is trained, allowing the effect of the refiner to be evaluated independently. All models are trained with $K=3$ refinement iterations and evaluated at $K=5$.

\begin{table*}[b!]
    \scriptsize
    \centering
    \caption{Ablation on voxel resolution $D$, visual feature injection, and the 2D convolution variant for refinement. Results are averaged across evaluation datasets. In this ablation experiment, all variants are built upon a frozen MoGe-2 (ViT-L) base model and are evaluated with 5 refinement iterations. $\star$~indicates our default setting.}
    \vspace{-4pt}
    \begin{tabular}{l|cc|cc|c|cc|cc|cc}
    \specialrule{.12em}{0em}{0em}

    \multirow{3}{*}{\textbf{Method}}
    & \multicolumn{4}{c|}{Local}
    & \multicolumn{1}{c|}{\!\!\!Boundary\!\!\!}
    & \multicolumn{6}{c}{Global} \\

    \cline{2-5} \cline{6-6} \cline{7-12}

    & \multicolumn{2}{c|}{\scriptsize Depth}
    & \multicolumn{2}{c|}{\scriptsize Point}
    & \multirow{2}{*}{\scriptsize F1$\uparrow$}
    & \multicolumn{2}{c|}{\scriptsize Relative Depth}
    & \multicolumn{2}{c|}{\scriptsize Relative Point}
    & \multicolumn{2}{c}{\scriptsize Metric Depth} \\

    \cline{2-5} \cline{7-12}

    & \scriptsize Rel\textsuperscript{d}$\downarrow$
    & \scriptsize $\delta_{0.01}^\text{d}\uparrow$
    & \scriptsize Rel\textsuperscript{p}$\downarrow$
    & \scriptsize $\delta_{0.01}^\text{p}\uparrow$
    &
    & \scriptsize Rel\textsuperscript{d}$\downarrow$
    & \scriptsize $\delta_{1}^\text{d}\uparrow$
    & \scriptsize Rel\textsuperscript{p}$\downarrow$
    & \scriptsize $\delta_{1}^\text{p}\uparrow$
    & \scriptsize Rel\textsuperscript{d}$\downarrow$
    & \scriptsize $\delta_{1}^\text{d}\uparrow$ \\
        \hline
        Base model & \cellcolor[rgb]{0.927,1.000,0.780} 3.81 & \cellcolor[rgb]{1.000,0.868,0.780} 47.1 & \cellcolor[rgb]{1.000,0.780,0.780} 3.19 & \cellcolor[rgb]{1.000,0.780,0.780} 46.6 & \cellcolor[rgb]{0.868,1.000,0.780} 15.6 & \cellcolor[rgb]{0.853,1.000,0.780} 5.98 & \cellcolor[rgb]{0.890,1.000,0.780} 94.1 & \cellcolor[rgb]{1.000,0.853,0.780} 8.73 & \cellcolor[rgb]{1.000,0.868,0.780} 90.4 & \cellcolor[rgb]{0.780,1.000,0.780} 15.6 & \cellcolor[rgb]{0.890,1.000,0.780} 77.3 \\
        2D Conv. & \cellcolor[rgb]{1.000,0.853,0.780} 4.09 & \cellcolor[rgb]{1.000,0.780,0.780} 45.0 & \cellcolor[rgb]{1.000,0.780,0.780} 3.19 & \cellcolor[rgb]{1.000,0.853,0.780} 47.0 & \cellcolor[rgb]{1.000,0.868,0.780} 14.8 & \cellcolor[rgb]{1.000,0.853,0.780} 6.15 & \cellcolor[rgb]{1.000,0.890,0.780} 93.8 & \cellcolor[rgb]{1.000,1.000,0.780} 8.54 & \cellcolor[rgb]{0.956,1.000,0.780} 91.1 & \cellcolor[rgb]{1.000,0.868,0.780} 16.4 & \cellcolor[rgb]{1.000,0.780,0.780} 76.2 \\
        RGB Input & \cellcolor[rgb]{1.000,0.927,0.780} 3.90 & \cellcolor[rgb]{1.000,0.956,0.780} 48.1 & \cellcolor[rgb]{1.000,0.868,0.780} 3.13 & \cellcolor[rgb]{1.000,0.927,0.780} 49.7 & \cellcolor[rgb]{1.000,0.956,0.780} 14.9 & \cellcolor[rgb]{1.000,1.000,0.780} 6.01 & \cellcolor[rgb]{1.000,1.000,0.780} 93.9 & \cellcolor[rgb]{1.000,0.927,0.780} 8.58 & \cellcolor[rgb]{1.000,0.956,0.780} 90.7 & \cellcolor[rgb]{1.000,0.956,0.780} 16.3 & \cellcolor[rgb]{1.000,0.780,0.780} 76.2 \\
        $D=100$ & \cellcolor[rgb]{1.000,1.000,0.780} 3.86 & \cellcolor[rgb]{0.868,1.000,0.780} 49.8 & \cellcolor[rgb]{0.868,1.000,0.780} 2.86 & \cellcolor[rgb]{0.853,1.000,0.780} 53.2 & \cellcolor[rgb]{0.780,1.000,0.780} 15.7 & \cellcolor[rgb]{0.927,1.000,0.780} 5.99 & \cellcolor[rgb]{0.890,1.000,0.780} 94.1 & \cellcolor[rgb]{0.780,1.000,0.780} 8.29 & \cellcolor[rgb]{0.780,1.000,0.780} 91.4 & \cellcolor[rgb]{0.956,1.000,0.780} 16.0 & \cellcolor[rgb]{1.000,1.000,0.780} 77.1 \\
        $D=200$~$\star$ & \cellcolor[rgb]{0.853,1.000,0.780} 3.70 & \cellcolor[rgb]{0.780,1.000,0.780} 51.4 & \cellcolor[rgb]{0.780,1.000,0.780} 2.76 & \cellcolor[rgb]{0.780,1.000,0.780} 55.1 & \cellcolor[rgb]{0.868,1.000,0.780} 15.6 & \cellcolor[rgb]{0.780,1.000,0.780} 5.88 & \cellcolor[rgb]{0.780,1.000,0.780} 94.2 & \cellcolor[rgb]{0.853,1.000,0.780} 8.34 & \cellcolor[rgb]{0.868,1.000,0.780} 91.2 & \cellcolor[rgb]{1.000,0.868,0.780} 16.4 & \cellcolor[rgb]{1.000,0.780,0.780} 76.2 \\
        $D=300$ & \cellcolor[rgb]{0.780,1.000,0.780} 3.64 & \cellcolor[rgb]{0.956,1.000,0.780} 48.8 & \cellcolor[rgb]{0.956,1.000,0.780} 2.92 & \cellcolor[rgb]{0.927,1.000,0.780} 52.0 & \cellcolor[rgb]{0.956,1.000,0.780} 15.1 & \cellcolor[rgb]{1.000,0.927,0.780} 6.11 & \cellcolor[rgb]{1.000,1.000,0.780} 93.9 & \cellcolor[rgb]{0.927,1.000,0.780} 8.36 & \cellcolor[rgb]{0.780,1.000,0.780} 91.4 & \cellcolor[rgb]{0.868,1.000,0.780} 15.8 & \cellcolor[rgb]{0.780,1.000,0.780} 77.7 \\
        $D=400$ & \cellcolor[rgb]{1.000,0.780,0.780} 4.26 & \cellcolor[rgb]{0.956,1.000,0.780} 48.8 & \cellcolor[rgb]{1.000,0.956,0.780} 3.11 & \cellcolor[rgb]{1.000,1.000,0.780} 50.4 & \cellcolor[rgb]{1.000,0.780,0.780} 14.3 & \cellcolor[rgb]{1.000,0.780,0.780} 7.12 & \cellcolor[rgb]{1.000,0.780,0.780} 91.4 & \cellcolor[rgb]{1.000,0.780,0.780} 9.65 & \cellcolor[rgb]{1.000,0.780,0.780} 88.3 & \cellcolor[rgb]{1.000,0.780,0.780} 16.5 & \cellcolor[rgb]{1.000,0.890,0.780} 76.3 \\
        \specialrule{.12em}{0em}{0em}
    \end{tabular}
    \vspace{-10pt}
    \label{tab:ablation}
\end{table*}

\vspace{-6pt}
\paragraph{Effect of voxel resolution $D$.}
The voxel resolution $D$ controls the spatial granularity of the 3D sparse convolutions: larger $D$ yields finer voxels, but also increases memory usage and sparse occupancy.
Increasing the voxel resolution from $D{=}100$ to moderate resolutions ($D{=}200$--$300$) substantially improves local metrics while preserving global accuracy, indicating that finer voxelization better captures fine geometric structures. However, pushing the resolution further to $D{=}400$ degrades both global and local performance due to increasingly sparse occupancy, which limits the effectiveness of the 3D sparse convolutions. We therefore adopt $D{=}200$ in the full model as the best trade-off between geometric precision, occupancy density, and efficiency.

\vspace{-6pt}
\paragraph{Effect of visual feature injection.}
To examine whether SSR benefits from high-level visual representations, we construct an RGB-input variant. This model removes DINO feature injection at the bottleneck. Instead, each voxel is initialized by concatenating the RGB value of its corresponding pixel to the voxel features. The refiner therefore still observes raw color and local appearance discontinuities, but no longer has direct access to the semantic and global-context information encoded by DINO features.
As shown in Table~\ref{tab:ablation}, the RGB-input variant improves local threshold accuracy over the frozen base model, but still remains clearly behind the default DINO-feature-injected refiner. Since this variant retains self-guided voxelization and sparse 3D convolutions, its improvement may come from both RGB appearance cues and the 3D refinement structure. These results indicate that SSR can improve geometry without DINO features to some extent, but the semantic and contextual information provided by visual features is crucial for resolving fine structures and disambiguating local 3D geometry.

\vspace{-6pt}
\paragraph{3D vs.\ 2D refinement.}
\begin{figure*}
    \centering
    \includegraphics[width=1\linewidth]{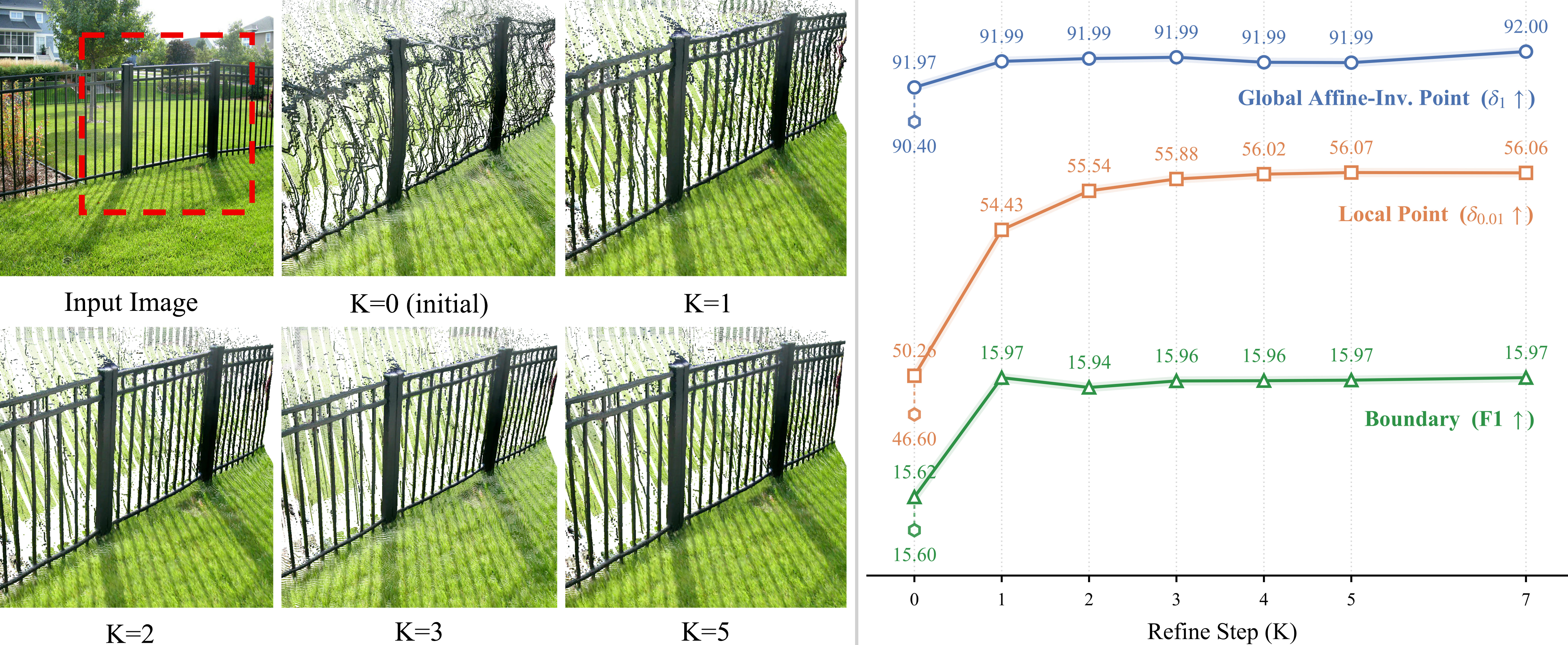}
    \vspace{-5pt}
    \caption{Effect of refinement iterations. Left: point map visualizations demonstrate progressively sharper reconstruction of thin structures. Right: quantitative metrics including global $\delta_{1}^{p}$, local $\delta_{0.01}^{p}$, and boundary F1 remain stable or improve with increasing refinement iterations. The model is trained with $K{=}3$. Larger $K$ values evaluate generalization beyond the training setting. The hexagon marker at $K{=}0$ denotes the performance of the base model MoGe-2.}
    \label{fig:ablation_refine_steps}
    \vspace{-10pt}
\end{figure*}
To evaluate the benefit of the introduced 3D inductive bias, we replace the sparse 3D U-Net with a parameter-matched 2D U-Net operating on the dense image grid while predicting the same log-depth residual $\Delta\zeta^{(k)}$. The only difference is whether refinement is performed on the dense 2D image grid or the self-guided sparse 3D voxel set $\mathcal{V}^{(k)}$.
The 2D variant in Table~\ref{tab:ablation} shows that simply increasing model capacity in 2D does not provide the gains achieved by 3D refinement. Despite being parameter-matched to the sparse 3D U-Net, the 2D variant yields only marginal changes on global metrics and fails to improve local geometry, with local point and depth metrics remaining comparable to or worse than the base model. This highlights a key limitation of 2D refinement: neighboring pixels in the image may correspond to geometrically distant surfaces in 3D, especially near occlusion boundaries. In contrast, sparse 3D convolutions separate such regions into distinct voxels, reducing cross-boundary mixing and enabling substantially sharper local geometry.

\vspace{-6pt}
\paragraph{Effect of refinement iterations.}
\label{sec:exp-iters}
We study the refinement behavior by varying the number of iterations at inference time, using $K \in \{0,1,\dots,5,7\}$ with the default jointly trained ViT-L model.
Figure~\ref{fig:ablation_refine_steps} shows how global $\delta_{1}^{p}$, local $\delta_{0.01}^{p}$, and boundary F1 evolve with $K$. 
At $K{=}0$ (\emph{no refinement}), the jointly trained base model already outperforms MoGe-2 baseline, demonstrating the effectiveness of joint training. 
As $K$ increases, local metrics improve substantially after the first step and continue to improve gradually before saturating around $K{=}5$, while global metrics remain nearly unchanged. This confirms that SSR selectively sharpens thin structures without disturbing overall geometry. Boundary F1 saturates even earlier, with most gains within the first iteration.
Although the refiner is trained with only $K{=}3$ iterations, performance does not degrade when applying more iterations at test time. Running up to $K{=}7$ yields slightly improved or stable metrics across all categories, indicating that the learned residual updates generalize well beyond the training horizon. This robustness suggests that each refinement step learns a stable, convergent correction rather than overfitting to a fixed number of iterations.
The qualitative example in Figure~\ref{fig:ablation_refine_steps} illustrates refinement behavior on thin fence railings, where SSR progressively separates occluding and background structures, recovering correct local geometry.
\section{Conclusion}
In this paper, we point out a potential architectural mismatch in existing monocular geometry estimation methods, where inherently 3D geometry is decoded within a 2D parameterization, leading to structural distortions and over-smoothed reconstructions around fine details and depth discontinuities. 
To address this problem, we propose a new paradigm that lifts geometry modeling from 2D image space to 3D space through a Self-Guided Sparse 3D Refiner (SSR), which performs geometry refinement directly on sparse 3D scene manifolds using 3D-aware feature aggregation. Extensive experiments demonstrate that MoGe-3 significantly improves local geometric fidelity while preserving globally plausible geometry. We believe our work significantly advances fine-detail monocular geometry estimation by presenting the first monocular approach that produces point maps both visually plausible in 2D depth maps and faithful in 3D geometry.

\paragraph{Limitations.} Our method is a regression-based paradigm, which inherently struggles to resolve pixel ambiguity along object boundaries. This limitation precludes the generation of perfectly sharp depth edges and introduces ``fly-points''. 
This could be addressed by integrating generative formulations such as~\cite{xu2025pixelperfectdepthsemanticsprompteddiffusion} into our framework.

\bibliographystyle{plainnat}
\bibliography{references}

\appendix
\clearpage
\setcounter{page}{1}
\renewcommand{\thetable}{\thesection.\arabic{table}}
\renewcommand{\thefigure}{\thesection.\arabic{figure}}
\setcounter{table}{0}
\setcounter{figure}{0}

\appendix

\section{Training Details}
\label{sec:app-training}

\subsection{Detailed Loss Design}
\label{sec:app-losses}

This appendix expands the main-text losses of Sec.~\ref{sec:losses}
with the exact solver formulations, weight clipping, and hyper-parameter
values used in our implementation.
\paragraph{Global affine-invariant loss.}
Following \cite{wang2025mogeunlockingaccuratemonocular}, supervision is invariant under the
optical-axis affine transform of Sec.~\ref{sec:preliminaries}. We solve the optimal $(s, t_z)$ alignment in closed form on a low-resolution copy
of the prediction and apply a depth-relative $\ell_1$ residual,
\begin{equation}
\mathcal{L}_{\mathrm{global}}^{(k)}
\;=\;
\left\langle\, w_{ij}\,
\left\|s^{(k)} \boldsymbol{p}^{(k)}_{ij} + t_z^{(k)} \boldsymbol{e}_z
                 - \widehat{\boldsymbol{p}}_{ij}\right\|_1 \,\right\rangle_{(i,j)\in\mathcal M},
\qquad
w_{ij} \propto 1/\widehat{Z}_{ij},
\label{eq:loss-global}
\end{equation}
where $\langle\cdot\rangle_{\mathcal M}$ denotes the mask-weighted mean
over valid pixels, and $(s^{(k)}, t_z^{(k)})$, which is the closed-form
weighted-median solution of
\begin{equation}
\big(s^{(k)},\, t_z^{(k)}\big) =
\arg\min_{s > 0,\, t_z}
\sum_{ij} w_{ij}
\big\| s\, \boldsymbol{p}^{(k)}_{ij} + t_z\, \boldsymbol{e}_z
       - \widehat{\boldsymbol{p}}_{ij} \big\|_1.
\label{eq:app-scale-shift}
\end{equation}

\paragraph{Radial-partition local loss.}
To penalize local geometric distortions that global alignment might overlook, we adopt the local loss from MoGe with two modifications:

(1) \textbf{Stochastic multi-scale radial partitioning}: At each loss evaluation, with equal probability we choose either the ground-truth point map or the current globally aligned prediction as the reference point map for partitioning:
\begin{equation}
\boldsymbol{r}_{ij}^{(k)} =
\begin{cases}
\widehat{\boldsymbol{p}}_{ij}, & \text{w.p. } 0.5,\\
\widetilde{\boldsymbol{p}}_{ij}^{(k)}, & \text{w.p. } 0.5,
\end{cases}
\qquad
\widetilde{\boldsymbol{p}}_{ij}^{(k)} = s^{(k)}\boldsymbol{p}_{ij}^{(k)} + t_z^{(k)}\boldsymbol{e}_z.
\end{equation}
We then group pixels into 3D local regions $\mathcal{G}_{\alpha,g}^{(k)}$ by applying the radial partition function to $\boldsymbol{r}_{ij}^{(k)}$, which quantizes both the projection direction and the log-distance in a scale-invariant manner:
\begin{equation}
    \mathcal G_{\alpha,g}^{(k)}=\left\{(i,j)\middle|\boldsymbol{\pi}_\alpha^{(k)}(i,j)=g\right\},\qquad \boldsymbol{\pi}_\alpha^{(k)}(i,j) = \left\lfloor \alpha \cdot
\left(\frac{\boldsymbol{r}_{ij}^{(k)}}{\|\boldsymbol{r}_{ij}^{(k)}\|_\infty},\log\|\boldsymbol{r}_{ij}^{(k)}\|_2 \right)\right\rceil,
\end{equation}
where $\alpha$ controls partition granularity and $g$ identifies a local partition under such granularity. We use $\alpha \in \{4, 16, 64\}$ to provide multi-scale supervision. In practice, we apply a random rotation and scale before the partition function for perturbation to the partition boundaries. The stochastic source selection encourages the partitions to respect both ground-truth geometry and the model's current foreground--background separation, reducing over-smoothing across local depth discontinuities.

(2) \textbf{Scale-shared local alignment}: For each local region $\mathcal{G}_{\alpha,g}^{(k)}$, we optimize a 3-DoF shift $\boldsymbol{t}_{\alpha,g}^{(k)}$ while sharing the global scale factor $s^{(k)}$. This design is motivated by the observation that local scale-shift alignments typically yield scale factors that deviate minimally from the global one. By fixing the scale, we simplify the objective to:
\begin{equation}
\mathcal{L}_{\mathrm{local}}^{(k)}=
\sum_{\alpha\in \{4,16,64\}}\left\langle\, w_{ij}\,
\left\|
s^{(k)} \boldsymbol{p}^{(k)}_{ij} + \boldsymbol{t}_{\alpha,g}^{(k)}
- \widehat{\boldsymbol{p}}_{ij} \right\|_1
\,\right\rangle_{(i,j) \in \mathcal{M}},\qquad g=\boldsymbol{\pi}_\alpha^{(k)}(i,j)
\label{eq:loss-local}
\end{equation}
where $\boldsymbol t_{\alpha,g}^{(k)}$ minimizes the alignment objective within its local region $\mathcal G_{\alpha,g}^{(k)}$:
\begin{equation}
\boldsymbol{t}_{\alpha,g}^{(k)} =
\arg\min_{\boldsymbol{t}}
\sum_{(i,j)\in\mathcal{G}_{\alpha,g}^{(k)}} w_{ij}
\big\| s^{(k)} \boldsymbol{p}^{(k)}_{ij} + \boldsymbol{t}
       - \widehat{\boldsymbol{p}}_{ij} \big\|_1,
\end{equation}

Compared to the original formulation, this approach improves the efficiency to linear computational time without compromising the effectiveness of local geometric supervision.

\paragraph{Edge-angle loss.} The Edge-angle loss can be written as:
\label{sec:app-loss-edge}

\begin{equation}
\mathcal{L}_{\mathrm{edge}}^{(k)} =
\frac{1}{\min(H, W)|\mathcal{D}_d|}
\sum_{(i,j) \in \mathcal{D}_d}\left(\angle\left(\partial_x \boldsymbol{p}^{(k)}_{ij},\partial_x \widehat{\boldsymbol{p}}_{ij}\right) + \angle\left(\partial_y \boldsymbol{p}^{(k)}_{ij},\partial_y \widehat{\boldsymbol{p}}_{ij}\right)\right),
\end{equation}
where $\mathcal{D}_d$ are the pixels whose two $d$-neighbors are both valid.

\paragraph{Auxiliary head losses.} The losses for the auxiliary heads are defined as:
\label{sec:app-loss-aux}

\begin{equation}
\mathcal{L}_{\mathrm{mask}} = \mathrm{BCE}(\boldsymbol{M}, \boldsymbol{\widehat{M}}),\quad
\mathcal{L}_{\mathrm{nml}} = \big\langle \angle(\boldsymbol{N},\widehat{\boldsymbol{N}})^{2} \big\rangle,\quad 
\mathcal{L}_{\mathrm{scale}} = \mathbb{I}(\hat{s}>0)\,
(\log s - \log \hat{s})^{2}.
\end{equation}

\subsection{Detailed Training Setup}
We train our model using loss weights
$(\lambda_{\mathrm{g}}, \lambda_{\ell}, \lambda_{\mathrm{e}}, \lambda_{\mathrm{m}}, \lambda_{\mathrm{n}}, \lambda_{\mathrm{s}})
= (1.0, 1.0, 1.0, 0.1, 0.1, 0.1)$.
The local term $\lambda_{\ell}$ is implemented as a multi-scale radial-partition loss evaluated at scales $\alpha \in \{4, 16, 64\}$, with equal weight $1.0$ at each scale.
Optimization is performed with AdamW ($\beta_1=0.9$, $\beta_2=0.999$, weight decay $10^{-2}$), together with gradient clipping at $\|g\|_2 \leq 1.0$.
We adopt a staged learning-rate schedule with warm-up and exponential decay. The backbone is frozen for the first $1{,}000$ steps, then linearly warmed up until step $2{,}000$. After warm-up, the backbone and 2D heads decay every $25{,}000$ steps, while the refiner decays every $10{,}000$ steps. Peak learning rates are $2\times10^{-4}$ for the refiner, $1\times10^{-4}$ for the 2D heads, and $5\times10^{-6}$ for the DINO backbone.
Training uses a global batch size of $96$ on $16 \times \mathrm{A100}$ GPUs with mixed precision. The DINO encoder operates in \texttt{bfloat16}, while all other modules use \texttt{float32}. For stability, the refiner is detached from the backbone for the first $5{,}000$ steps and then jointly optimized with full gradient flow.

\subsection{Training Data}
We train the full model on a large, heterogeneous mixture of synthetic datasets
with pixel-perfect geometry, including 
3D Ken Burns~\cite{niklaus20193dkenburnseffect}, 
ApolloSynthetic~\cite{apollosynthetic}, 
EDEN~\cite{le21wacv}, 
GTA-SfM~\cite{Wang2019gtasfm}, 
Hypersim~\cite{roberts2021hypersimphotorealisticsyntheticdataset}, 
IRS~\cite{wang2021irslargenaturalisticindoor}, 
MatrixCity~\cite{li2023matrixcitylargescalecitydataset}, 
MidAir~\cite{Fonder2019MidAir}, 
MVS-Synth~\cite{huang2018mvsynth}, 
Objaverse~\cite{deitke2022objaverseuniverseannotated3d}, 
OmniWorld~\cite{zhou2025omniworldmultidomainmultimodaldataset}, 
Structured3D~\cite{zheng2020structured3dlargephotorealisticdataset}, 
Synscapes~\cite{Synscapes}, 
Synthia~\cite{Ros2016synthia}, 
TartanAir~\cite{wang2020tartanairdatasetpushlimits}, 
UnrealStereo4K~\cite{Tosi2021SMDNetsUnrealStereo4K}, and 
UrbanSyn~\cite{G_mez_2025}, and real / sensor-reconstructed datasets including 
A2D2~\cite{geyer2020a2d2}, 
ARKitScenes~\cite{dehghan2021arkitscenes}, 
Argoverse2~\cite{Argoverse2}, 
BlendedMVS~\cite{yao2020blendedmvslargescaledatasetgeneralized}, 
MegaDepth~\cite{li2018megadepthlearningsingleviewdepth}, 
ScanNet++~\cite{yeshwanthliu2023scannetpp}, 
Taskonomy~\cite{zamir2018taskonomydisentanglingtasktransfer}, 
and 
Waymo~\cite{sun2020waymo}
 that supply real-world
appearance and sensor statistics.

\section{Details of Local Metric Evaluation }
\subsection{Local Fine-Grained Mask Construction}
\label{sec:appendix_fine_grained_mask}

We design an automatic pipeline to identify fine-grained regions in depth maps, such as thin structures, depth discontinuities, and delicate geometric details. The pipeline consists of two stages: coarse mask detection and segment-aware expansion. 

\paragraph{Coarse mask construction.}
We combine two complementary detectors: (1) \textit{Multi-scale Laplacian residual detector.} \ \  We first compute the residual $r_s(p) = |d(p) - \tilde{d}_s(p)|$ between the disparity and its downsampled-then-upsampled version under different scales $s \in \{8, 16, 32\}$. A pixel is classified as belonging to a fine-grained region when $r_s(p) > 3.0 \cdot \hat{\sigma}_s$, where $\hat{\sigma}_s$ is the robust standard deviation based on Median Absolute Deviation (MAD). (2) \textit{Morphological top-hat / bottom-hat.} \ \  For structuring element sizes $k \in \{3, 5, 9, 17\}$, the white top-hat $d - \gamma(d)$ detects thin \emph{near} structures (e.g., poles), and the black top-hat $\phi(d) - d$ detects thin \emph{far} gaps (e.g., inter-foliage holes). Each residual is thresholded at $3.0$ robust standard deviations.

The final coarse mask is obtained by taking the union over both detectors across all scales. The robust scale $\hat{\sigma}$ is estimated using the Median Absolute Deviation (MAD): $\hat{\sigma} = 1.4826 \cdot \mathrm{median}(|r - \mathrm{median}(r)|)$, with a fallback to the above-median tail when the MAD is zero.

\paragraph{Segment-aware expansion.}
We employ SAM2~\cite{ravi2024sam2segmentimages} to partition the RGB image into distinct regions. For each segment $S_i$, we compute its support density with respect to the coarse structure mask $\mathcal{S}$ as
\begin{equation}
\rho_i = \frac{|S_i \cap \mathcal{S}|}{|S_i|}.
\end{equation}
A segment is retained only if it satisfies the following criteria:
\begin{equation}
\rho_i \geq 0.3,\qquad
|S_i \cap \mathcal{S}| \geq 5,\qquad
|S_i| \leq 0.05HW,
\end{equation}
thereby preserving only small-to-medium regions with sufficient structural evidence. The final mask is obtained by taking the union of all selected segments and intersecting it with the valid depth region.

\subsection{Evaluation Protocol}
\label{sec:appendix_fine_grained_eval}

Given the automatically generated fine-grained masks (Sec.~\ref{sec:appendix_fine_grained_mask}), we design a per-segment local evaluation protocol that measures how well a model reconstructs delicate geometric structures, independent of global alignment errors.

\paragraph{Per-segment shift with global scale.}
We first fit a single global scale $s^*$ on all valid pixels using weighted least-squares:
\begin{equation}
    s^* = \arg\min_s \sum_{p \in \text{valid}} w_p \left\| s \cdot \hat{z}_p - z_p^{\text{gt}} \right\|^2,
\end{equation}
where $w_p = 1/z_p^{\text{gt}}$ for depth and $w_p = 1/\|\boldsymbol{x}_p^{\text{gt}}\|$ for 3D points. Then, for each SAM2 segment $S_i$ (with $|S_i| \geq 10$ pixels) within the fine-grained mask, we fit a per-segment shift $t_i$ while keeping the global scale fixed:
\begin{equation}
    t_i = \arg\min_t \sum_{p \in S_i} w_p \left\| s^* \cdot \hat{z}_p + t - z_p^{\text{gt}} \right\|^2.
\end{equation}
For 3D point maps, the shift $\boldsymbol{t}_i \in \mathbb{R}^3$ is a per-segment translation vector. The locally-aligned prediction is $\tilde{z}_p = s^* \hat{z}_p + t_i$ for depth, or $\tilde{\boldsymbol{x}}_p = s^* \hat{\boldsymbol{x}}_p + \boldsymbol{t}_i$ for points. We adopt this formulation because it avoids both the scale-collapse degeneracy of per-segment affine alignment and the sensitivity to field-of-view mismatch inherent in scale-only alignment, while remaining applicable to both depth maps and point maps in a unified manner.

\paragraph{Metrics and Aggregation.}
On the locally-aligned prediction within each segment, we compute AbsRel and $\delta_{0.01}$. Final results are reported as the average across all segments.

\section{Additional Quantitative Results}
\paragraph{Evaluation results on individual datasets}
\label{sec:appendix_more_results}
We report detailed per-dataset results for global metrics, local metrics, and boundary accuracy in Table~\ref{tab:appendix_global}, Table~\ref{tab:appendix_local} and Table~\ref{tab:appendix_boundary}, respectively.
\begin{table*}[ht!]
\centering
\begingroup
\scriptsize
\setlength{\tabcolsep}{1.2pt}
\begin{tabular}{l|cc|cc|cc|cc|cc|cc|cc|cc|cc|ccc}
\toprule
Method & \multicolumn{2}{c|}{NYUv2} & \multicolumn{2}{c|}{KITTI} & \multicolumn{2}{c|}{ETH3D} & \multicolumn{2}{c|}{DIODE} & \multicolumn{2}{c|}{HAMMER} & \multicolumn{2}{c|}{iBims-1} & \multicolumn{2}{c|}{Sintel} & \multicolumn{2}{c|}{Spring} & \multicolumn{2}{c|}{Synth4K} & \multicolumn{3}{c}{Avg.} \\
\cmidrule{2-3}\cmidrule{4-5}\cmidrule{6-7}\cmidrule{8-9}\cmidrule{10-11}\cmidrule{12-13}\cmidrule{14-15}\cmidrule{16-17}\cmidrule{18-19}\cmidrule{20-22}
 & Rel$\downarrow$ & $\delta_1\uparrow$ & Rel$\downarrow$ & $\delta_1\uparrow$ & Rel$\downarrow$ & $\delta_1\uparrow$ & Rel$\downarrow$ & $\delta_1\uparrow$ & Rel$\downarrow$ & $\delta_1\uparrow$ & Rel$\downarrow$ & $\delta_1\uparrow$ & Rel$\downarrow$ & $\delta_1\uparrow$ & Rel$\downarrow$ & $\delta_1\uparrow$ & Rel$\downarrow$ & $\delta_1\uparrow$ & Rel$\downarrow$ & $\delta_1\uparrow$ & Rank$\downarrow$ \\
\midrule
\multicolumn{22}{c}{Affine-invariant depth map} \\
\midrule
Depth Pro & \cellcolor[rgb]{1.000,0.829,0.780} 3.67 & \cellcolor[rgb]{1.000,0.890,0.780} 98.2 & \cellcolor[rgb]{1.000,0.878,0.780} 5.12 & \cellcolor[rgb]{1.000,0.927,0.780} 96.8 & \cellcolor[rgb]{1.000,0.780,0.780} 4.97 & \cellcolor[rgb]{1.000,0.843,0.780} 96.4 & \cellcolor[rgb]{1.000,0.878,0.780} 4.66 & \cellcolor[rgb]{1.000,0.890,0.780} 95.6 & \cellcolor[rgb]{1.000,0.835,0.780} 3.30 & \cellcolor[rgb]{0.780,1.000,0.780} 99.6 & \cellcolor[rgb]{1.000,0.780,0.780} 3.23 & \cellcolor[rgb]{1.000,0.853,0.780} 98.3 & \cellcolor[rgb]{1.000,0.835,0.780} 15.8 & \cellcolor[rgb]{1.000,0.835,0.780} 80.1 & \cellcolor[rgb]{1.000,0.829,0.780} 21.7 & \cellcolor[rgb]{1.000,0.829,0.780} 70.4 & \cellcolor[rgb]{1.000,0.843,0.780} 10.4 & \cellcolor[rgb]{1.000,0.829,0.780} 86.6 & \cellcolor[rgb]{1.000,0.829,0.780} 8.10 & \cellcolor[rgb]{1.000,0.843,0.780} 91.3 & \cellcolor[rgb]{1.000,0.829,0.780} 8.28 \\
\bottomrule
\end{tabular}
\endgroup
\caption{Comparison on global metrics across datasets and metric groups. DA3 refers to Depth Anything 3, and PPD refers to Pixel-Perfect Depth.}
\label{tab:appendix_global}
\end{table*}

\begin{table*}[ht!]
\centering
\caption{Comparison on local metrics across datasets and metric groups.}
\resizebox{\textwidth}{!}{%
\begingroup
\scriptsize
\setlength{\tabcolsep}{1.2pt}
\begin{tabular}{l|cc|cc|ccc|cc|cc|ccc}
\toprule
\multirow{3}{*}{Method} & \multicolumn{7}{c|}{Local fine-grained point map} & \multicolumn{7}{c}{Local fine-grained depth map} \\
 & \multicolumn{2}{c}{Spring} & \multicolumn{2}{c}{Synth4K} & \multicolumn{3}{c|}{Avg.} & \multicolumn{2}{c}{Spring} & \multicolumn{2}{c}{Synth4K} & \multicolumn{3}{c}{Avg.} \\
 & Rel$\downarrow$ & $\delta_{0.01}\uparrow$ & Rel$\downarrow$ & $\delta_{0.01}\uparrow$ & Rel$\downarrow$ & $\delta_{0.01}\uparrow$ & Rank$\downarrow$ & Rel$\downarrow$ & $\delta_{0.01}\uparrow$ & Rel$\downarrow$ & $\delta_{0.01}\uparrow$ & Rel$\downarrow$ & $\delta_{0.01}\uparrow$ & Rank$\downarrow$ \\
\midrule
Depth Pro & \cellcolor[rgb]{0.780,1.000,0.780} 2.12 & \cellcolor[rgb]{0.890,1.000,0.780} 56.1 & \cellcolor[rgb]{0.976,1.000,0.780} 3.90 & \cellcolor[rgb]{1.000,0.829,0.780} 34.3 & \cellcolor[rgb]{0.927,1.000,0.780} 3.01 & \cellcolor[rgb]{0.976,1.000,0.780} 45.2 & \cellcolor[rgb]{0.945,1.000,0.780} 4.50 & \cellcolor[rgb]{0.976,1.000,0.780} 3.72 & \cellcolor[rgb]{0.927,1.000,0.780} 52.0 & \cellcolor[rgb]{0.976,1.000,0.780} 4.57 & \cellcolor[rgb]{1.000,0.878,0.780} 41.7 & \cellcolor[rgb]{0.976,1.000,0.780} 4.15 & \cellcolor[rgb]{0.976,1.000,0.780} 46.9 & \cellcolor[rgb]{0.976,1.000,0.780} 5.50 \\
\bottomrule
\end{tabular}
\endgroup
}
\label{tab:appendix_local}
\end{table*}

\begin{table}[ht!]
\small
\centering
\caption{Boundary F1 score (radius=1) on iBims-1, HAMMER, Sintel, and Spring.}
\begin{tabular}{lcccccc}
\toprule
Method & iBims-1 & HAMMER & Sintel & Spring & Avg. F1$\uparrow$ & Avg.\ Rk.$\downarrow$ \\
\midrule
Depth Pro & \cellcolor[rgb]{1.000,0.976,0.780} 14.5 & \cellcolor[rgb]{1.000,0.976,0.780} 5.07 & \cellcolor[rgb]{0.945,1.000,0.780} 34.8 & \cellcolor[rgb]{0.829,1.000,0.780} 10.9 & \cellcolor[rgb]{0.890,1.000,0.780} 16.3 & \cellcolor[rgb]{0.906,1.000,0.780} 4.75 \\
UniDepth V2 & \cellcolor[rgb]{1.000,0.829,0.780} 11.3 & \cellcolor[rgb]{1.000,0.878,0.780} 4.66 & \cellcolor[rgb]{0.835,1.000,0.780} 35.9 & \cellcolor[rgb]{1.000,0.829,0.780} 6.26 & \cellcolor[rgb]{1.000,0.890,0.780} 14.5 & \cellcolor[rgb]{1.000,0.843,0.780} 7.00 \\
DA 3 & \cellcolor[rgb]{1.000,0.780,0.780} 8.31 & \cellcolor[rgb]{0.976,1.000,0.780} 5.36 & \cellcolor[rgb]{1.000,0.780,0.780} 9.14 & \cellcolor[rgb]{1.000,0.780,0.780} 3.61 & \cellcolor[rgb]{1.000,0.780,0.780} 6.61 & \cellcolor[rgb]{1.000,0.780,0.780} 8.75 \\
UniK3D & \cellcolor[rgb]{1.000,0.878,0.780} 12.8 & \cellcolor[rgb]{1.000,0.829,0.780} 4.65 & \cellcolor[rgb]{0.835,1.000,0.780} 35.9 & \cellcolor[rgb]{1.000,0.927,0.780} 6.97 & \cellcolor[rgb]{1.000,0.945,0.780} 15.1 & \cellcolor[rgb]{1.000,0.906,0.780} 6.50 \\
PPD & \cellcolor[rgb]{1.000,0.927,0.780} 13.8 & \cellcolor[rgb]{0.780,1.000,0.780} 7.80 & \cellcolor[rgb]{1.000,0.835,0.780} 20.7 & \cellcolor[rgb]{0.927,1.000,0.780} 8.61 & \cellcolor[rgb]{1.000,0.835,0.780} 12.7 & \cellcolor[rgb]{0.969,1.000,0.780} 5.25 \\
InfiniDepth & \cellcolor[rgb]{0.878,1.000,0.780} 17.9 & \cellcolor[rgb]{0.829,1.000,0.780} 7.50 & \cellcolor[rgb]{0.780,1.000,0.780} 39.8 & \cellcolor[rgb]{0.780,1.000,0.780} 11.8 & \cellcolor[rgb]{0.780,1.000,0.780} 19.3 & \cellcolor[rgb]{0.780,1.000,0.780} 1.75 \\
MoGe-2 & \cellcolor[rgb]{0.976,1.000,0.780} 17.6 & \cellcolor[rgb]{1.000,0.780,0.780} 4.54 & \cellcolor[rgb]{1.000,0.890,0.780} 32.0 & \cellcolor[rgb]{0.976,1.000,0.780} 8.22 & \cellcolor[rgb]{1.000,1.000,0.780} 15.6 & \cellcolor[rgb]{1.000,0.843,0.780} 7.00 \\
\bottomrule
\end{tabular}
\label{tab:appendix_boundary}
\end{table}

\paragraph{Runtime analysis}
We report the runtime statistics in Table~\ref{tab:appendix_runtime} to characterize the practical computational cost of each method. All measurements are obtained on a single NVIDIA A100 GPU with a forward batch size of 1, and every reported latency value is the average over 200 sampled inputs. For both Pixel-Perfect Depth and InfiniDepth, we do not employ the prompt-based metric-depth branch, and only time the relative-depth prediction stage. For all other models, we record the complete forward pass that produces both the point map and the depth map.

\begin{table}[ht!]
\small
\centering
\caption{Runtime statistics measured on a single NVIDIA A100 GPU for single-frame inference. MoGe-2 (ViT-G) is our scaled-up MoGe-2-style base model trained following the MoGe-2 configuration.}
\label{tab:appendix_runtime}
\begin{tabular}{lccccc}

\toprule
\multirow{2}{*}{Method} & \multirow{2}{*}{Native Resolution} & \multicolumn{2}{c}{Latency (ms)} \\
\cmidrule(lr){3-4}
 & & FP16 & FP32  \\
\midrule
\multirow{3}{*}{MoGe-2 (ViT-L)}
 & $484^2$ & 29 & 82   \\
 & $700^2$ & 39 & 157  \\
 & $840^2$ & 55 & 238  \\
\midrule
\multirow{3}{*}{MoGe-2 (ViT-G)}
 & $484^2$ & 64 & 239   \\
 & $700^2$ & 98 & 486  \\
 & $840^2$ & 137 & 724  \\
\midrule
\multirow{2}{*}{Depth Anything 3}
 & $378{\times}504$  & 152 & 305 & \\
 & $630{\times}840$  & 213 & 737 & \\
\midrule
Depth Pro
 & $1536^2$ & 142 & 910  \\
\midrule
\multirow{2}{*}{InfiniDepth}
 & $512{\times}672$  & 133 & 343  \\
 & $768{\times}1024$ & 258 & 724  \\
\midrule
\multirow{2}{*}{Pixel-Perfect Depth}
 & $512{\times}672$  & 275 & 1054  \\
 & $768{\times}1024$ & 598 & 3123  \\
\midrule
\multirow{3}{*}{\textit{Ours (ViT-L) 3-step}}
 & $484^2$  & 89 & 149  \\
 & $700^2$  & 121 & 260  \\
 & $840^2$  & 154 & 370  \\
\midrule
\multirow{3}{*}{\textit{Ours (ViT-G) 3-step}}
 & $484^2$  & 115 & 308  \\
 & $700^2$  & 177 & 591  \\
 & $840^2$  & 234 & 860  \\
\bottomrule
\end{tabular}

\end{table}

\section{Additional Qualitative Results}
\label{sec:appendix_visual}

\begin{figure}[t]
    \centering
    \includegraphics[width=\linewidth]{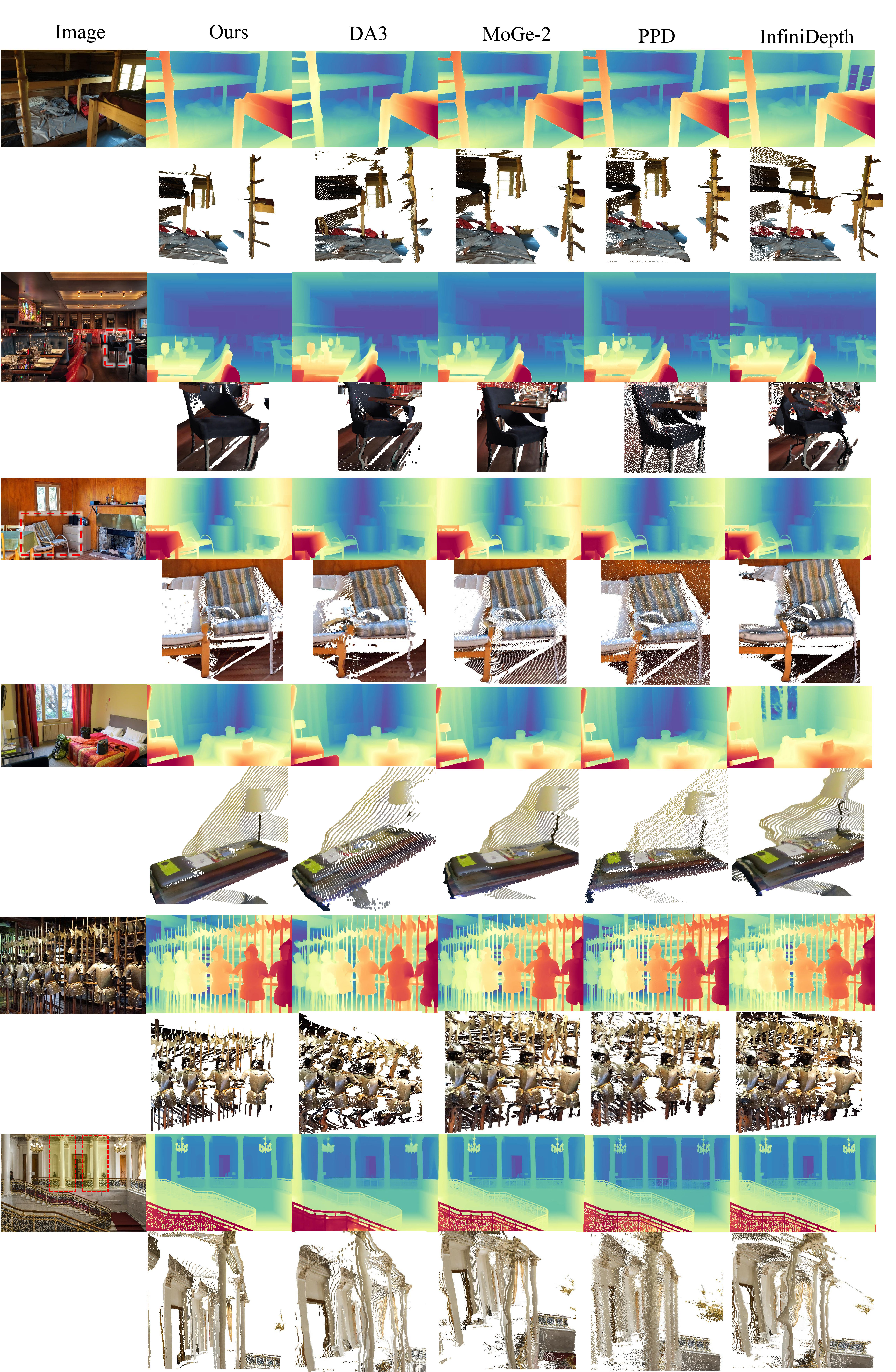}
    \caption{\textbf{Additional qualitative results and comparisons.} Our method preserves fine structures and sharp boundaries across diverse scenes.}
    \label{fig:samples1}
\end{figure}

\begin{figure}[t]
    \centering
    \includegraphics[width=\linewidth]{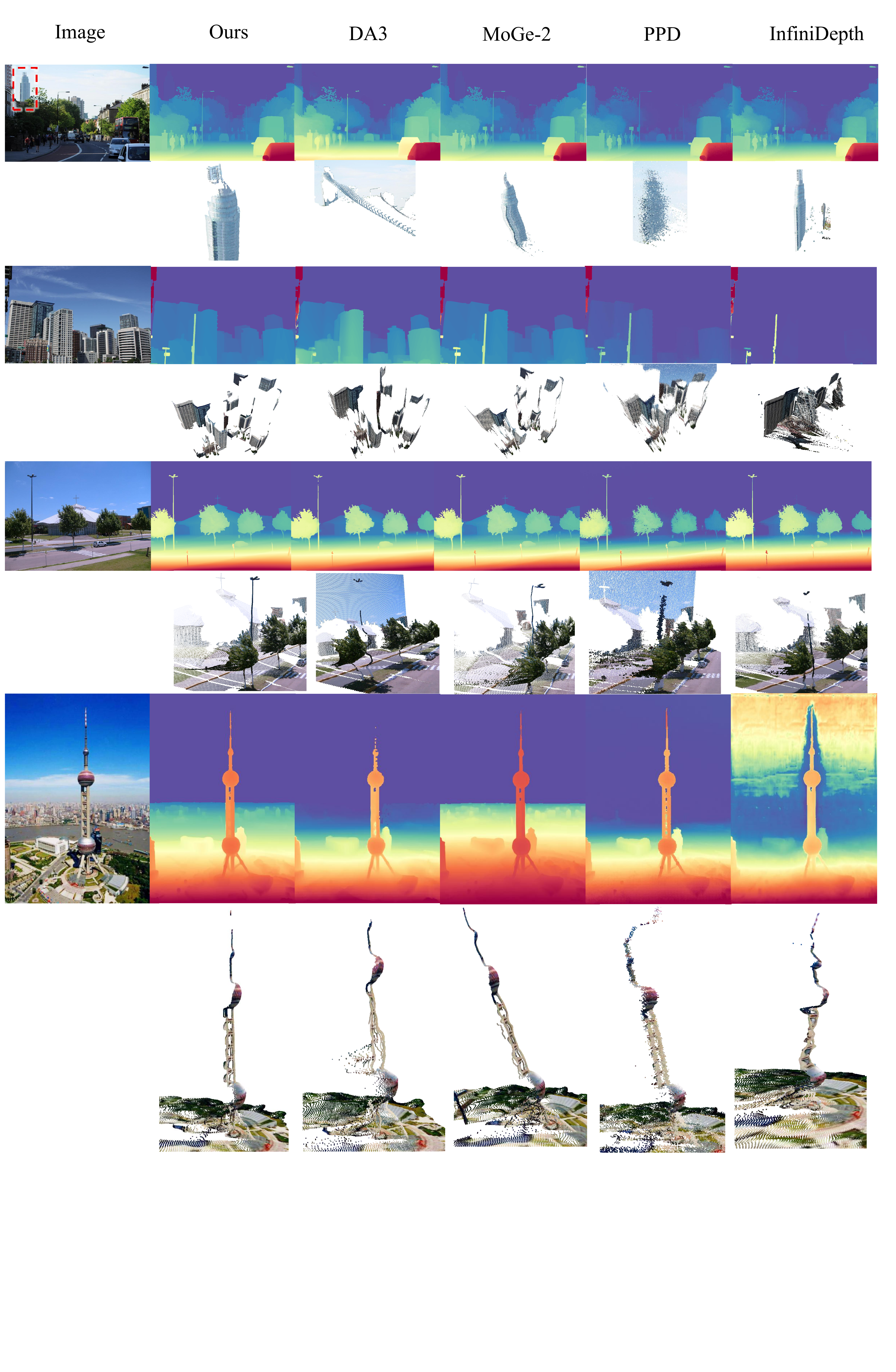}
    \captionsetup{skip=-3cm}
    \caption{\textbf{Additional qualitative results.} More examples demonstrating high-fidelity geometric reconstruction and robustness to challenging regions.}
    \label{fig:samples2}
\end{figure}

We provide additional qualitative results on diverse scenes in Figure \ref{fig:samples1} and \ref{fig:samples2}, highlighting fine-grained structures, depth discontinuities, and geometric details.

\section{Examples of Local Fine-Grained Mask}
\begin{figure}[t]
    \centering
    \includegraphics[width=\linewidth]{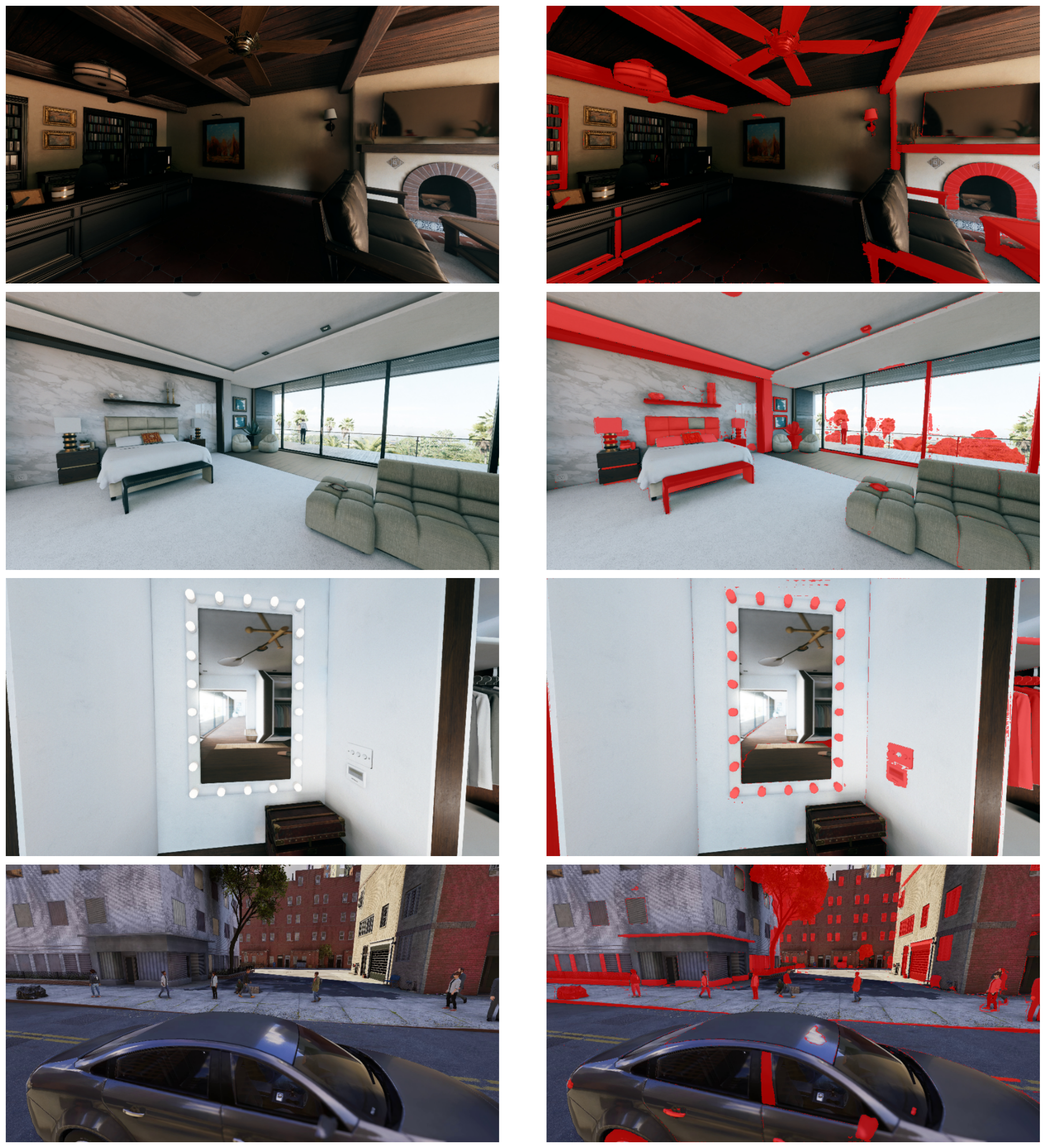}
    \caption{\textbf{Local fine-grained masks.} For each example, we show the input image (left) and the corresponding
local fine-grained mask overlaid in red (right). These masks highlight
thin structures and high-frequency regions used for evaluation.}
    \label{fig:synth4k_masks}
\end{figure}

Figure~\ref{fig:synth4k_masks} illustrates several examples of our generated local fine-grained masks from the Synth4K dataset.

\end{document}